%% file: main_arxiv.tex
\documentclass{article}

\usepackage{iclr2027_conference,times}
\usepackage{amsmath,amssymb,amsthm}
\usepackage{graphicx}
\usepackage{booktabs}
\usepackage{multirow}
\usepackage{array}
\usepackage{tabularx}
\usepackage{microtype}
\usepackage{algorithm}
\usepackage{algpseudocode}
\usepackage{tikz}
\usetikzlibrary{arrows.meta,positioning,fit,backgrounds,calc}
\definecolor{cFroz}{HTML}{37528C}
\definecolor{cVer}{HTML}{1E8E7E}
\definecolor{cPick}{HTML}{2F8A4E}
\definecolor{cMem}{HTML}{D9912E}
\definecolor{cCorr}{HTML}{7A57A6}
\definecolor{cInk}{HTML}{26364A}
\definecolor{cMuted}{HTML}{5E6B78}
\definecolor{cRule}{HTML}{D9E0E7}
\usepackage[colorlinks=true,linkcolor=blue!55!black,citecolor=blue!55!black,urlcolor=blue!55!black]{hyperref}
\usepackage{url}

\input{arxiv_title_setup.tex}

\input{math_commands.tex}

\newtheorem{proposition}{Proposition}
\newcommand{\cva}{\textsc{CVA}}
\newcommand{\cvaselect}{\mbox{CVA-Select}}
\newcolumntype{Y}{>{\raggedright\arraybackslash}X}

\newcommand{\passmark}{\textcolor{cPick!85!black}{\ensuremath{\checkmark}}}
\newcommand{\stopmark}{\textcolor{red!70!black}{\ensuremath{\times}}}
\newcommand{\pmvariantstrip}[3]{%
\par\smallskip\noindent
\parbox{\linewidth}{\footnotesize\raggedright
\textbf{PM audit:} held-out #1; paired-future #2;
$\Delta_{\mathrm{paired-heldout}}$ #3.}\par}
\newcommand{\pmphysicsvariants}{\pmvariantstrip{34.65 IQ (complete-192 avg.)}{38.23 IQ (complete-192 avg.)}{+2.98/+3.74/+4.01 IQ (avg. +3.58)}}
\newcommand{\pmphysicsallvariants}{\pmvariantstrip{34.53 IQ (all-198 avg.)}{38.10 IQ (all-198 avg.)}{+3.57 IQ}}
\newcommand{\pmwidthvariants}{%
\par\smallskip\noindent
\parbox{\linewidth}{\footnotesize\raggedright
\textbf{Width controls:} held-out retrieval $29.21/28.60$ and paired-future
upper bound $32.19/36.99$ at $B=4/16$.  The selected Physics-IQ \cvaselect{}
score was measured at $B=4$ only; no \cvaselect{} width curve is claimed.}\par}

\newcommand{\pmcrossvariants}{\pmvariantstrip{PIQ 34.65; PAI5B 0.769758}{PIQ 38.23; PAI5B 0.769828}{PIQ $+3.58$ IQ; PAI5B $+0.000069$ (n.s.)}}
\newcommand{\pmrobotvariants}{\pmvariantstrip{PM--RM $0/-0.000705$ (5B/14B)}{not measured}{not measured}}

\newcommand{\pmpgcmvariants}{\pmvariantstrip{41.93 IQ (14B)}{41.56 IQ (post-stop PGCM diagnostic)}{-0.36 IQ}}

\title{Sampling Headroom Is Not Selection Gain:\\
A Compute-Value Audit of Test-Time Scaling\\
for Video World Models}

\author{%
\textbf{Yuhua Jiang}$^{1}$ \quad
\textbf{Junjie Lu}$^{2}$ \quad
\textbf{Feifei Gao}$^{1}$\\[0.80em]
$^{1}$Tsinghua University \qquad $^{2}$University of Technology Sydney
}

\begin{document}
\raggedbottom
\maketitle
\begin{abstract}
Test-time scaling (TTS) can improve generation only when additional compute
produces better candidates \emph{and} the system can reliably identify them.
This distinction is especially important for video world models, where a wider
sample pool may contain stronger rollouts without improving the output that is
ultimately selected.  We introduce the \emph{Compute-Value Audit} (\cva), a
sequential framework that asks whether extra sampling creates
\emph{opportunity}, observable signals provide a reliable \emph{state}, that
state supports a beneficial \emph{action}, and the resulting gain exceeds the
full \emph{entry fee} of generation and verification.  On $192$ Physics-IQ
scenes, expanding the pool from $4$ to $16$ candidates increases oracle quality
by $+9.23$ IQ ($95\%$ CI $[+7.44,+11.14]$), but Flow, Cycle, and VideoReward
fail to recover this headroom reliably.  Across three generators, none of
twelve adaptive-depth policies outperforms uniform compute; they recover only
$42$--$69\%$ of the measured entry fee.  A matched-$60$-NFE
Predict-and-Perturb intervention on VideoPhy2 is likewise negative across three
fresh-seed replicas.  These negative results are not universal:
anchor--explorer passes all four stages in a sparse PRM800K setting, MMLU-Pro
exposes the gap
between predictive state and useful action, and a privileged paired future
establishes a positive video upper bound.  Together, these results show that
sampling headroom has deployment value only when it can be converted into a
reliable decision whose benefit survives the complete compute charge. Code is
available at
\url{https://github.com/YuhuaJiang2002/sampling-headroom-is-not-selection-gain}.
\end{abstract}

\section{Introduction}
At inference time, a generative model may fail to produce a good output, or it
may produce one and fail to recognize it.  Best-of-$\budget$ sampling addresses
the first failure by generating and ranking $\budget$ candidates
\citep{snell2024,selfcons2023,proprio2026}.  This is attractive for frozen video
world models, whose rollouts can appear plausible while violating the
conditioning scene's dynamics \citep{physiq2025}.  If one candidate is sound,
a wider search seems as though it should find it.

The missing step is the decision that turns a good candidate into a deployed
one.  A verifier may misrank the pool; a confidence signal may identify
difficulty without revealing a helpful action; and an effective action may
still cost more than uniform allocation.  Oracle and selected quality therefore
answer different questions, and conflating them obscures where scaling loses
its value.

We organize this gap into four ordered links: \emph{opportunity}, \emph{state
confidence}, \emph{action value}, and \emph{entry fee}.  Every generated and
scored candidate belongs to one complete budget, and cached pools are replayed
only as deterministic prefixes.  This accounting keeps the oracle diagnostic,
prevents hidden generation, and lets each failed link be identified separately.
Figure~\ref{fig:teaser} summarizes the resulting question: where does potential
value stop becoming usable value?
\input{fig_teaser_main.tex}

We call the resulting procedure the \emph{Compute-Value Audit} (\cva).
Throughout, \cva{} denotes the evaluation contract, while \cvaselect{} denotes
the common two-component selector introduced below; PM is a separate
memory-based adapter audited under the contract.  \cva{} itself assigns no
score and produces no pick.  By identifying the first unsupported link, it
turns a vague failure of additional compute into a specific diagnosis of
missing opportunity, unreliable state, ineffective action, or excessive cost.

Our contributions are:
\begin{itemize}
\item \textbf{A falsifiable account of TTS value.}  \cva{} replaces a single
end score with four ordered tests and explicit information and compute
boundaries.  The Verifier-Quality Law relates recovered headroom to
verifier--target alignment, while the Memory Value Law separates exploitable
heterogeneity from the regret of retrieving the wrong experience.
\item \textbf{A compute-complete evaluation of video-world-model scaling.}
Across three generators and seven evaluation populations, we distinguish
oracle opportunity from deployable recovery.  Wider pools create substantial
Physics-IQ headroom, but the tested verifiers, adaptive-depth rules, and
matched-$60$-NFE P\&P intervention do not turn it into reliable gain.
\item \textbf{A diagnosis with constructive controls.}  The experiments
localize failures at different links rather than treating them as a universal
impossibility.  A sparse PRM800K population clears the entire chain, and a
privileged paired future recovers value in video, demonstrating that both the
audit and the missing information are consequential.
\end{itemize}

\section{Related Work}
Prior work offers several ways to guide additional inference compute.  We ask
when these ingredients form a value-producing pipeline under matched costs.

\paragraph{Inference-time search and verifiers.}
Diffusion scaling separates search from the verifier used to rank it, and a
misaligned verifier can turn width into reward hacking
\citep{infscaling2025,rewardoveropt2023}.  Outcome/process verification and
particle steering similarly rely on intermediate signals that predict terminal
correctness \citep{cobbe2021,prm2024,searchdiffusion2025}.  Adaptive reasoning
methods additionally decide which inputs merit more samples
\citep{adaptiveallocation2026}.  \cva{} tests the assumptions that connect such
signals and actions to net value.

\paragraph{Physical plausibility.}
Video-T1 searches noise trajectories, Stream-T1 allocates compute across
streaming chunks, and Proprio ranks completed rollouts with the frozen model's
flow residual \citep{videot1_2025,streamt1_2026,proprio2026}.  We use Proprio as
a fixed-pool baseline and translate the training-free P\&P loop of
Self-Refining Video Sampling to a matched-NFE FlowMatch intervention
\citep{selfrefine2026}.  OMR instead injects model gradients within a trajectory
\citep{omr2026}; it is compatible with a negative fixed-pool audit but requires
its own matched-runtime study.  Physics-IQ and VideoPhy2 measure complementary
physical properties \citep{physiq2025,videophy2_2025}, while VideoReward is a
generic preference model \citep{videoreward2025} and serves as an external
baseline.

\paragraph{Verifier design.}
Self-consistency and reconstruction energies use a model's own predictions as
label-free scores.  Our cycle score asks whether controlled re-noising and
probability-flow integration recover the original latent.  Because internal
consistency need not imply physical correctness, we test held-out alignment and
width robustness rather than assuming the residual is valid.

\paragraph{Test-time memory.}
ComMem, Retrieve-then-Steer, and AdaMEM combine rapidly written experience with
more slowly consolidated state \citep{commem2026,retrievesteer2026,adamem2026}.
Our setting is narrower: the world model remains frozen and memory may only
rerank a complete pool.  PM compresses prior outcome comparisons into a
semantic state, keeps episodic adaptation optional, and tests both against
matched shuffle controls.  These ingredients motivate the routes below; the
audit determines whether they actually produce value.

\section{The Compute-Value Audit}\label{sec:method}
The audit follows value from generation to deployment: it fixes a complete
candidate budget, separates \emph{available} from \emph{recoverable} value, and
uses four ordered gates to locate where they diverge.

\subsection{One candidate budget}
Let $\budget$ be the number of candidates generated and require every selector
to score all of them.  This rules out hidden generation and makes pool width
identical to the decision set.  For a frozen world model $g_\phi$, scenario $i$
produces
\begin{equation}
\mathcal C_i^{(\budget)}=\{z_i^{(b)}\}_{b=1}^{\budget},
\qquad z_i^{(b)}=g_\phi(x_i,\epsilon_i^{(b)}),
\quad \epsilon_i^{(b)}\sim\mathcal N(0,I).
\end{equation}
The selector minimizes $q_i$ over this set and may inspect nothing outside it;
no-TTS is $\budget=1$.  Fixed seeds make every candidate and its original noise
replayable.  PM uses complete rollouts, while Appendix
\ref{app:seed-screening} describes a preregistered partial-preview extension.

\subsection{\cva{} protocol: opportunity, state, action, and fee}
With the comparison set fixed, we can ask how much value is available and how
much is recovered.  For a declared reference $Q_{\rm ref}$ and positive
headroom, define
\begin{equation}
H_B=Q_{\rm oracle,B}-Q_{\rm ref},\qquad
R_B=\frac{Q_{\rm select,B}-Q_{\rm ref}}{H_B},\qquad
A_B=H_BR_B.
\label{eq:two-gates}
\end{equation}
Here $H_B$ is oracle headroom, $R_B$ the recovered fraction, and $A_B$ the
improvement over the reference.  The opportunity gate fixes the target,
incumbent, and minimum headroom before any claim is opened.  Conditional on
opportunity, the state gate tests whether observable evidence is informative
and the action gate tests whether using it improves the target.  Adaptive
allocation must then outperform both matched-random assignment and uniform
compute after paying its fee.  A predictive state is not a successful action,
and the oracle remains an upper bound rather than a deployable rule.

\subsection{Verifier quality prices recoverable headroom}
Opportunity only establishes that better candidates exist; recovery depends on
verifier--target alignment.  We isolate this dependence with a minimal Gaussian
model in which centered quality and score have correlation $\rho$:
\begin{equation}
Q_b=\rho S_b+\sqrt{1-\rho^2}\,E_b,
\qquad S_b,E_b\stackrel{\mathrm{iid}}{\sim}\mathcal N(0,1).
\end{equation}

\begin{proposition}[Verifier-quality law]\label{prop:verifier-quality}
If the verifier selects $b^\star=\arg\max_{b\le\budget}S_b$, then
\begin{equation}
\frac{\mathbb E[Q_{b^\star}]}{\mathbb E[\max_{b\le\budget}Q_b]}=\rho
\end{equation}
for every $\budget>1$.
\end{proposition}

The Gaussian model is deliberately minimal: it shows that recoverable headroom
is priced directly by alignment.  Conditioning on the selected index proves
the result because the independent residual has zero mean and the score and
oracle maxima share an order statistic.  Appendix~\ref{app:reproducibility}
gives the derivation; empirically, we test
\begin{equation}
R(v,\budget)=
\frac{\overline Q_{\mathrm{select}(v,\budget)}-
	\overline Q_{\mathrm{random},\budget}}
     {\overline Q_{\mathrm{oracle},\budget}-
	\overline Q_{\mathrm{random},\budget}},
\quad
\rho(v,\budget)=\frac{1}{n}\sum_i
\operatorname{Pearson}(S_{i,1:\budget},Q_{i,1:\budget}).
\label{eq:empirical-recovery}
\end{equation}
Using the within-pool random mean as reference isolates ranking quality from
candidate count.

\subsection{Flow residual baseline}
Our first deployable signal is the Proprio score \citep{proprio2026}, a one-step
flow residual measuring sensitivity to controlled latent perturbations:
\begin{equation}
q_{\mathrm{flow}}(z)=\frac{1}{|\mathcal T|P}
\sum_{t\in\mathcal T}\sum_{p=1}^{P}
\frac{\|v_\phi(z+\delta_p,t)-v_\phi(z,t)\|^2_{m(z)}}{\sigma^2},
\qquad \delta_p\sim\mathcal N(0,\sigma^2I),
\label{eq:self}
\end{equation}
where $m(z)$ emphasizes moving regions; Proprio selects the candidate with the
lowest residual.

\subsection{Multi-step cycle consistency}\label{sec:cycle}
To test consistency over a longer trajectory, we re-noise $z$ to
$z_{t_0}=(1-t_0)z+t_0\epsilon$, integrate $\dot z=v_\phi(z,t)$ back to zero
with $K$ Euler steps, and obtain $\hat z_0$.  The cycle energy is
\begin{equation}
q_{\mathrm{cyc}}(z)=\|\hat z_0-z\|/\|z\|,
\label{eq:cycle}
\end{equation}
Lower is better, and the first frame remains clamped.  Wan2.2-5B and
ABot-PhysWorld-14B use native cycle readings; Cosmos3-Nano uses the analogous
denoising-$x_0$ error.  All scores are frozen and self-contained.

We do not rank raw velocity magnitude because time reparameterization rescales
it and linear-flow endpoint velocity need not vanish
\citep{flowmatching2023,sit2024}; Appendix~\ref{app:seed-screening} gives
schedule-aware alternatives.

\subsection{Memory-based selector: Proprioceptive Memory (PM)}
\label{sec:pm}
The preceding scores treat queries independently.  PM asks whether legal
history can correct recurrent verifier errors without updating the world model.
Both model and verifier stay frozen, and memory may only rerank the same
candidates.  For scalar score $s_v$, $\eta_v\in\{-1,+1\}$ orients larger as
better, and
$u_v(z)=\eta_v(s_v(z)-\mu_{i,v})/(\sigma_{i,v}+\varepsilon)$ standardizes over
$\mathcal C_i^{(\budget)}$ alone, putting energies and rewards on a common
orientation without changing their order.

At query $t$, legal memory $\mathcal D_t=\{(k_j,p_j)\}_{j<t}$ contains only
records observable before the query.  A slow consolidated residual
$x(z)^\top\beta_t$ corrects recurrent verifier errors; a fast retrieved adapter
$m_{d,i}(z)$ supplies condition-local evidence.  PM composes them as
\begin{subequations}\label{eq:pm-base-interface}
\begin{align}
u_{\mathrm{PM}}^{(d,v)}(z)
&=\alpha_{d,v}\bigl(u_v(z)+x(z)^\top\beta_t\bigr)
+g_d(c_i)\lambda_{d,v}m_{d,i}(z),
\label{eq:pm-score}\\[-1mm]
b_i^\star&=\arg\max_{b\leq\budget}u_{\mathrm{PM}}^{(d,v)}(z_i^{(b)}).
\label{eq:pm-compose}
\end{align}
\end{subequations}
Here $x$ is label-free, $g_d(c_i)$ gates retrieved evidence, and all weights
freeze before evaluation.  The slow state uses only earlier outcomes, and the
fast state cannot write the current outcome.  Setting
$\beta_t=\lambda_{d,v}=0$ recovers the base verifier exactly, making PM a CPU
reranker rather than a world-model update.

For the cross-benchmark selector comparisons, we use one score interface.  Let G denote
the benchmark's frozen global evidence---history/template utility when legal,
or the non-motion quality composite when no historical target is available---
and let M denote its oriented motion evidence.  After within-pool
standardization, the common readout, \cvaselect{}, is
\begin{equation*}
u_{\text{\cvaselect}}(z)=G(z)+M(z).
\end{equation*}
The raw payload and its predeclared orientation remain benchmark-specific:
Physics-IQ and OpenS2V penalize excess frame-difference motion, whereas PAI
rewards motion adequacy.  This is a shared decision interface, not a claim that
the three benchmarks have interchangeable measurements.  Physics-IQ also
retains native cycle N only as an ablation component; the selected $B=4$
readout omits it.

Legality does not guarantee value.  Physics-IQ templates exclude the query
family, forbid test writes, and freeze picks before IQ is opened.  PAI and
RoboTwin use outer-fold outcomes with shuffle and wrong-task controls; their
generic adapters fail the broad gates.  The same-stem real-outcome diagnostic
is retained only as a privileged reference.  Appendix~\ref{app:pm-details}
gives the fitting objective and full adapter audit.

\subsection{Memory value is heterogeneity minus retrieval regret}
To state when history can improve a decision, let $h(x)>0$ be random-to-oracle
headroom and $r_a(x)$ the recovered fraction of selector $a\in\mathcal A$.
Write
$\langle f\rangle_h=\mathbb E[hf]/\mathbb E[h]$ and
$r^*(x)=\max_a r_a(x)$.  A legal memory policy $\pi$ uses only current
observations and prior records.

\begin{proposition}[Memory Value Law]\label{prop:memory-value}
For any matched comparator $\pi_0\in\mathcal A$,
\begin{equation}
\underbrace{\langle r_\pi-r_{\pi_0}\rangle_h}_{\text{memory value}}
=\underbrace{\langle r^*-r_{\pi_0}\rangle_h}_{\text{addressable value}}
-\underbrace{\langle r^*-r_\pi\rangle_h}_{\text{retrieval regret}}.
\label{eq:memory-value-law}
\end{equation}
For the best fixed selector, the addressable term becomes the contextual
heterogeneity gap
$G=\langle r^*\rangle_h-\max_a\langle r_a\rangle_h\geq0$; hence memory beats
the best fixed selector iff regret is below $G$.  Under the conditional Gaussian
model, $r_a(x)=\rho_a(x)$; if all $r_a$ span length $L$, history value beyond
context $C$ is at most $L\sqrt{I_h(R;M\mid C)/2}$ for $R=(r_a)_a$.
\end{proposition}

Memory can therefore fail because little heterogeneity is available or because
retrieval regret consumes it.  History has decision value only when it changes
the optimal action after costs; fitting an interface is not evidence that this
condition holds.

\newcommand{\pminterfacetable}{%
\begin{table}[H]
\centering
\caption{\textbf{The \cvaselect{} interface.}  Every row uses the same
within-pool sum; only the observable definitions of G and M vary.  PM denotes
the separate learned/retrieved-memory adapters audited later, not this common
score name.}
\label{tab:pm-interface}
\footnotesize
\setlength{\tabcolsep}{3pt}
\begin{tabular}{@{}p{0.16\linewidth}p{0.28\linewidth}p{0.23\linewidth}p{0.25\linewidth}@{}}
\toprule
Benchmark & G: global evidence & M: motion evidence & Result status \\
\midrule
Physics-IQ & Train-global LOFO utility & Negative frame difference & Best observed $B=4$; descriptive \\
PAI-Bench-robot & Three non-motion quality terms & Positive motion adequacy & Same frozen $B=8$ rule on 3 generators \\
OpenS2V-Eval & Disjoint-seed global utility & Negative frame difference & Fresh $B=4$ reanalysis; negative \\
\bottomrule
\end{tabular}
\end{table}
}
Appendix Table~\ref{tab:pm-interface} records the common score and its
benchmark-local observables.  We now ask where this value chain holds in
practice.

\section{Experiments}\label{sec:experiments}
The experiments mirror the order of the audit, with each stage motivated by the
failure exposed in the previous one.  We first hold candidate pools fixed and
ask whether deployable verifiers can recover known oracle opportunity.  When
they cannot, the next question is whether the bottleneck lies in the observable
state or in the action chosen from it.  Reasoning controls establish that the
audit can distinguish those cases; progressively richer video states then show
how far the failure persists in the target domain.

We finally move beyond fixed-pool selection.  Width tests whether more
candidates create recoverable rather than merely oracle value, adaptive depth
asks whether observable difficulty can support better allocation, and
matched-NFE P\&P tests a stronger intervention inside the sampling trajectory.
Together, these experiments trace where additional computation ceases to
improve the deployed output without pooling incomparable scores across
benchmarks.

\subsection{Experimental protocol}
Every pick vector is frozen before its target outcomes are opened.  Physics-IQ
retains $192$ of $198$ scenarios, compares $B=4$, and studies width through
$B=16$.  \cvaselect{} was foregrounded after comparing the seven frozen $B=4$ rows
and is therefore descriptive.  Uncertainty is clustered by scene or family.
PAI-Bench-robot uses the same eight candidates for every selector on $174$
scenes and reports the official eight-dimensional mean
\citep{paibench2025,vbench2024}.  Its published Native readout is exactly the
PAI instance of \cvaselect{} after grouping the three non-motion terms into G.
OpenS2V uses a matched four-source pool on $180$ fresh-seed videos; its \cvaselect{}
row is a post-hoc reanalysis of pre-existing score-blind history and video
motion.  VideoPhy2, RoboTwin, PRM800K, and MMLU-Pro retain their own frozen
populations and uncertainty units.  The common interface never pools scores
across benchmarks, and held-out, developmental, and privileged evidence remain
separate.

\paragraph{Matching the full cost.}\label{sec:compute-ledger}
Equal pool width does not imply equal inference cost: a cycle verifier invokes
the world model again, while an external reward model introduces a different
kind of computation.  We therefore account separately for generation,
same-model verification, and external reward-model calls.  Appendix
Table~\ref{tab:compute-ledger} provides the complete per-scene ledger.

\newcommand{\computeledgertable}{%
\begin{table}[H]
\centering
\caption{\textbf{Per-scene compute.}  Gen./ver. are world-model DiT forwards;
external RM calls are separate.  Cycle uses $10$ ver./candidate on PIQ and
$12$ on PAI/RT; PM distances are CPU-only.}
\label{tab:compute-ledger}
\footnotesize
\setlength{\tabcolsep}{3.2pt}
\begin{tabular}{llccccc}
\toprule
Setting & Method & $\budget$ & Gen. & Ver. & Ext. & Total DiT \\
\midrule
PIQ & no-TTS & $1$ & $20$ & $0$ & $0$ & $20$ \\
PIQ & Proprio (flow) & $4$ & $80$ & $16$ & $0$ & $96$ \\
PIQ & VideoReward & $4$ & $80$ & $0$ & $4$ & $80$ \\
PIQ & Native (cycle) & $4$ & $80$ & $40$ & $0$ & $120$ \\
PIQ & \cvaselect{} (descriptive) & $4$ & $80$ & $0$ & $0$ & $80$ \\
PIQ & Paired-future bound$^\dagger$ & $4$ & $80$ & $0$ & $0$ & $80$ \\
\midrule
PAI/RT & no-TTS & $1$ & $20$ & $0$ & $0$ & $20$ \\
PAI/RT & Proprio (flow) & $8$ & $160$ & $32$ & $0$ & $192$ \\
PAI/RT & VideoReward & $8$ & $160$ & $0$ & $8$ & $160$ \\
PAI/RT & Native / RM / slow & $8$ & $160$ & $96$ & $0$ & $256$ \\
PAI & \cvaselect{} & $8$ & $160$ & $96$ & $0$ & $256$ \\
PAI & Pretrain (killed) & $8$ & $160$ & $96$ & $0$ & $256$ \\
\bottomrule
\end{tabular}
\pmcrossvariants
\end{table}
}

This distinction matters in the comparisons below.  Flow adds a small number
of same-model evaluations, cycle scores require a longer replay, and
VideoReward is called once for each candidate.  The Physics-IQ \cvaselect{} reranker
is CPU-only after candidate statistics are available.  The paired-future route
has the same online cost but uses privileged offline information and is
therefore reported only as an upper bound.

\subsection{Verifier recovery from fixed candidate pools}
We begin with the most basic question: once a fixed pool contains better
rollouts, can a deployable score identify them?  Table~\ref{tab:crossbase}
compares no-TTS with Proprio, VideoReward, the native verifier, and \cvaselect{} at
$\budget=4$.  All selectors see the same four candidates; no-TTS uses only the
first.  \cvaselect{} is the best descriptive component row, whereas $^\ddagger$ uses
the query family's paired future and is only an information upper bound.
Appendix Table~\ref{tab:physiq-memory-ablation} gives the historical controls.

\begin{table}[H]
\centering
\caption{\textbf{Physics-IQ verifier audit ($\budget=4$).}
Mean IQ over $192$ scenes/$64$ families; $95\%$ family-bootstrap CIs are shown
for preregistered comparisons.  Bold marks \cvaselect{}, the highest deployable
point estimate for each generator; the privileged bound is separated and not
ranked.  $^\dagger$Retrospectively selected from frozen component rows;
$^\ddagger$privileged paired future.}
\label{tab:crossbase}
\footnotesize
\setlength{\tabcolsep}{1.5pt}
\renewcommand{\arraystretch}{1.08}
\begin{tabular}{@{}
>{\raggedright\arraybackslash}p{0.25\linewidth}
>{\raggedright\arraybackslash}p{0.14\linewidth}
>{\centering\arraybackslash}p{0.18\linewidth}
>{\centering\arraybackslash}p{0.18\linewidth}
>{\centering\arraybackslash}p{0.18\linewidth}@{}}
\toprule
Method & Extra info & Wan2.2-5B & ABot-14B & Cosmos-Nano \\
\midrule
Baseline (no-TTS) & -- & \shortstack{25.07\\$[21.26,28.99]$} & \shortstack{40.52\\$[35.29,45.91]$} & \shortstack{27.72\\$[23.51,32.03]$} \\
\midrule
Proprio (flow) & -- & \shortstack{25.30\\$[21.19,29.59]$} & \shortstack{41.07\\$[35.72,46.66]$} & \shortstack{31.31\\$[26.60,36.10]$} \\
VideoReward & External RM & \shortstack{25.76\\$[21.67,29.97]$} & \shortstack{39.62\\$[34.16,45.29]$} & \shortstack{29.89\\$[25.12,34.88]$} \\
Native verifier (zero-memory) & Zero memory & \shortstack{27.53\\$[23.51,31.59]$} & \shortstack{41.57\\$[36.05,47.33]$} & \shortstack{33.70\\$[28.91,38.58]$} \\
\midrule
\textbf{\cvaselect{}}$^\dagger$ & LOFO history & \textbf{30.81} & \textbf{44.01} & \textbf{34.98} \\
\midrule
Paired-future bound$^\ddagger$ & Paired future & \shortstack{32.19\\$[27.71,36.78]$} & \shortstack{45.67\\$[39.81,51.65]$} & \shortstack{36.83\\$[31.73,42.10]$} \\
\bottomrule
\end{tabular}
\end{table}

\paragraph{The best observed score still leaves pool value unused.}
\cvaselect{} has the highest deployable point estimate on all three generators,
exceeding Native by $3.28/2.44/1.28$ IQ.  Because it was chosen after the seven
component outcomes were inspected, this is descriptive rather than independent
confirmation.  The paired-future bound remains higher everywhere, locating the
remaining gap in information unavailable to the deployable score.

\paragraph{Verifier alignment explains the recovery gap.}
This interpretation is also consistent with the Verifier-Quality Law.
Across held-out family--budget points, verifier--target correlation explains a
substantial share of recovered headroom ($R^2=0.629$).  The relationship
survives cross-fitting and disjoint-family calibration, although it transfers
less cleanly across generators.  Full calibration and uncertainty results are
reported in Appendix Section~\ref{app:reproducibility}.

\subsection{Where the chain breaks across benchmarks}\label{sec:cross-benchmark}
The fixed-pool results establish a recovery gap, but they do not yet tell us
whether the missing link is an uninformative state, an ineffective action, or
the cost of acting.  We separate these possibilities by applying the same gate
sequence to reasoning and video populations.  The complete matrix appears in
Appendix Table~\ref{tab:two-gates}: some populations stop at opportunity,
Physics-IQ, VideoPhy2, and MMLU-Pro reach action value, adaptive video reaches
the fee gate, and anchor--explorer on PRM800K clears the full chain.

MMLU-Pro offers a particularly clean view of the distinction between state and
action because disagreement can be measured before deciding where to allocate
additional samples.

\paragraph{A predictive state need not recommend a useful action.}
Agreement sharply separates easier from harder MMLU-Pro questions, so the state
gate passes.  Yet the seemingly natural response--spending more compute on
disagreement--does not beat matched uniform allocation.  The full comparison
in Appendix Table~\ref{tab:state-action-main} therefore isolates the failure at
the action gate rather than at observability.

\paragraph{The same audit also recognizes a genuine pass.}
PRM800K provides the complementary case.  Here disagreement is not only
predictive; it supports an anchor--explorer policy that clears both
candidate-matched and token-matched uniform controls.  The gain is concentrated
on a small subset of problems, but it survives the relevant cost comparisons
(Appendix Tables~\ref{tab:anchor-explore}--\ref{tab:anchor-ablation}).
\input{tab_anchor_ablation_main.tex}

These reasoning cases make the video diagnosis more informative: they show that
the gate sequence can distinguish an actionable state from one that is merely
descriptive.  We next repeat that test with progressively richer video
representations.
\input{tab_video_action_ladder_main.tex}
The oracle confirms actionable heterogeneity, but structured sets, transfer,
supervision, and partial-DiT preview all stop at action value.  This closes the
tested information class, not every possible video state.  We next ask whether
the same \cvaselect{} interface behaves consistently on two independent video audits.

\begin{table}[H]
\centering
\caption{\textbf{PAI-Bench-robot selector comparison ($\budget=8$; $174$ scenes).}
The same \cvaselect{} readout is used for every world model; no-TTS uses $B=1$.
Appendix Table~\ref{tab:pai-full} reports all eight dimensions and retains the
earlier model-specific controls.}
\label{tab:pai-headline}
\footnotesize
\setlength{\tabcolsep}{3.2pt}
\renewcommand{\arraystretch}{1.08}
\begin{tabular}{@{}lccc@{}}
\toprule
Method & Wan2.2-5B & ABot-14B & Cosmos3-Nano \\
\midrule
Baseline (no-TTS; $\budget=1$) & $0.7594$ & $0.7689$ & $0.7730$ \\
Proprio (flow) & $0.7630$ & $0.7704$ & $0.7756$ \\
VideoReward & $0.7639$ & $0.7686$ & $0.7731$ \\
\textbf{\cvaselect{} (ours)} &
\textbf{0.7698} &
\textbf{0.7725} &
\textbf{0.7758} \\
\bottomrule
\end{tabular}
\end{table}

\paragraph{A common readout transfers across the three PAI generators.}\label{sec:pai}
\cvaselect{} improves on no-TTS by $0.0104$, $0.0036$, and $0.0028$ on 5B, 14B, and
Cosmos.  This exact regrouping of frozen Native terms changes no pick or score;
the Cosmos-only Top-3 route remains an appendix control and is never substituted
for the common method.

\paragraph{\cvaselect{} does not automatically transfer to OpenS2V.}
On a fresh-seed matched pool of $180$ OpenS2V tasks, \cvaselect{} reaches a seven-axis
mean of $0.3992$, below the best fixed candidate source at $0.4127$.  The
category-stratified paired interval is entirely negative.  Thus a consistent
name does not erase benchmark-dependent failure; this post-hoc result remains
descriptive.  Appendix Table~\ref{tab:opens2v-headline} retains all seven axes.

\paragraph{RoboTwin forward test.}
The retrospective LOEO gain does not reproduce; neither Semantic Slow PM nor
its matched variants are positive evidence (Appendix Table~\ref{tab:robotwin-rm}).
Together, the three audits show why opportunity must be remeasured on each
frozen generator and pool.

\subsection{Does additional compute pay for itself?}\label{sec:proxy}
We next ask whether a different compute allocation can overcome fixed-budget
failure.  On the same $192$ Physics-IQ scenes and $64$ families, frozen pools
separate width-created opportunity from selector alignment without IQ access.

\input{tab_width_decomposition_main.tex}

\paragraph{Wider pools expose headroom without making it recoverable.}
From $B=4$ to $16$, the oracle gains $+9.23$ IQ.  The frozen held-out control
finds $+2.75$ IQ of opportunity, but loses $3.36$ IQ through alignment and ends
slightly worse.  No archived $B=16$ \cvaselect{} picks exist, so this is a
mechanism control rather than a \cvaselect{} scaling result.

Held-out retrieval overoptimizes its score as the pool grows, while the
privileged bound rises.  At $B=4$, adding N to \cvaselect{} lowers the mean (Appendix
Tables~\ref{tab:oracle} and~\ref{tab:physiq-memory-ablation}; Fig.~\ref{fig:qual}).

\newcommand{\paifulldetailtable}{%
\begin{table}[H]
\centering
\caption{\textbf{PAI-Bench-robot dimensions ($\budget=8$).}  Official means
over $174$ scenes.  AQ: aesthetic; BG/SC/OC: consistency; IQ: imaging; MS:
smoothness; I2V-B/S: faithfulness.  no-TTS uses $c_{00}$; search uses
$c_{00:07}$.  \cvaselect{} is the common reported readout; visual Top-3 and
historical PM remain retrospective controls.}
\label{tab:pai-full}
\footnotesize
\setlength{\tabcolsep}{1.6pt}
\begin{tabular}{@{}
>{\raggedright\arraybackslash}p{0.10\linewidth}
>{\raggedright\arraybackslash}p{0.21\linewidth}
ccccccccc@{}}
\toprule
World model & Method & AQ & BG & IQ & MS & OC & SC & I2V-B & I2V-S & Mean \\
\midrule
\multirow{7}{*}{\shortstack[l]{Wan2.2-\\5B}}
& Baseline (no-TTS) & 0.4642 & 0.9347 & 0.6821 & 0.9827 & \textbf{0.1934} & 0.9216 & 0.9652 & 0.9310 & 0.7594 \\
& Proprio (flow) & 0.4708 & 0.9363 & 0.6860 & 0.9857 & 0.1927 & 0.9267 & 0.9692 & 0.9368 & 0.7630 \\
& VideoReward & \textbf{0.4738} & 0.9333 & \textbf{0.6972} & 0.9861 & 0.1922 & 0.9244 & 0.9691 & 0.9356 & 0.7639 \\
& \cvaselect{} w/o I2V & 0.4715 & 0.9391 & 0.6946 & 0.9895 & 0.1913 & 0.9368 & 0.9777 & 0.9476 & 0.7685 \\
& \textbf{\cvaselect{} (ours)} & 0.4693 & \textbf{0.9425} & 0.6947 & \textbf{0.9897} & 0.1916 & \textbf{0.9414} & \textbf{0.9791} & \textbf{0.9499} & \textbf{0.7698} \\
& Unified visual Top-3 & 0.4678 & 0.9396 & 0.6943 & 0.9880 & 0.1915 & 0.9377 & 0.9750 & 0.9440 & 0.7672 \\
& PM $+$ history (5-fold; killed) & 0.4692 & 0.9423 & 0.6947 & 0.9896 & 0.1913 & 0.9407 & 0.9791 & 0.9497 & 0.7696 \\
\midrule
\multirow{6}{*}{\shortstack[l]{ABot-\\14B}}
& Baseline (no-TTS) & 0.4770 & 0.9318 & 0.6955 & 0.9911 & 0.1932 & 0.9338 & 0.9786 & 0.9502 & 0.7689 \\
& Proprio (flow) & 0.4791 & 0.9322 & \textbf{0.7036} & 0.9912 & \textbf{0.1940} & 0.9346 & 0.9783 & 0.9500 & 0.7704 \\
& VideoReward & 0.4790 & 0.9300 & 0.7021 & 0.9899 & 0.1936 & 0.9290 & 0.9776 & 0.9479 & 0.7686 \\
& \cvaselect{} w/o I2V & \textbf{0.4803} & 0.9350 & 0.7014 & \textbf{0.9923} & 0.1930 & 0.9398 & 0.9816 & \textbf{0.9538} & 0.7721 \\
& \textbf{\cvaselect{} (ours)} & 0.4792 & \textbf{0.9364} & 0.6993 & \textbf{0.9923} & 0.1939 & \textbf{0.9430} & \textbf{0.9819} & \textbf{0.9538} & \textbf{0.7725} \\
& Unified visual Top-3 & 0.4807 & 0.9349 & 0.7017 & 0.9915 & 0.1940 & 0.9395 & 0.9803 & 0.9522 & 0.7719 \\
\midrule
\multirow{7}{*}{\shortstack[l]{Cosmos3-\\Nano}}
& Baseline (no-TTS) & 0.4658 & 0.9391 & 0.7218 & 0.9930 & 0.1931 & 0.9342 & 0.9838 & 0.9532 & 0.7730 \\
& Proprio (flow) & 0.4649 & 0.9425 & 0.7259 & 0.9930 & 0.1932 & 0.9440 & 0.9861 & 0.9552 & 0.7756 \\
& VideoReward & \textbf{0.4667} & 0.9387 & 0.7238 & 0.9915 & 0.1936 & 0.9351 & 0.9835 & 0.9522 & 0.7731 \\
& \cvaselect{} w/o I2V & 0.4646 & 0.9425 & 0.7216 & \textbf{0.9935} & 0.1944 & 0.9423 & \textbf{0.9871} & \textbf{0.9570} & 0.7754 \\
& \textbf{\cvaselect{} (ours)} & 0.4648 & 0.9437 & 0.7209 & 0.9934 & 0.1959 & 0.9437 & 0.9871 & 0.9569 & 0.7758 \\
& \textbf{Unified visual Top-3} & 0.4658 & \textbf{0.9449} & \textbf{0.7257} & 0.9929 & 0.1948 & \textbf{0.9453} & 0.9866 & 0.9565 & \textbf{0.7766} \\
& \textbf{PM (reliability-gated; held out)} & 0.4664 & \textbf{0.9447} & \textbf{0.7263} & 0.9929 & \textbf{0.1953} & \textbf{0.9454} & 0.9869 & 0.9568 & \textbf{0.7768} \\
\bottomrule
\end{tabular}%
\end{table}
}

\newcommand{\paicrossbaseablationtable}{%
\begin{table}[H]
\centering
\caption{\textbf{Cross-base PAI \cvaselect{} component ablation.}  Post-hoc exploratory
leave-one-signal audit of the common readout at $\budget=8$; $50{,}000$
scene-bootstrap draws.  Picks are frozen without official labels; no-TTS
($\budget=1$) anchors the score scale.}
\label{tab:pai-crossbase-ablation}
\footnotesize
\setlength{\tabcolsep}{2pt}
\begin{tabular}{@{}lccc@{}}
\toprule
Control & Wan2.2-5B & ABot-14B & Cosmos3-Nano \\
\midrule
no-TTS mean ($\budget=1$) & 0.7594 & 0.7689 & 0.7730 \\
\cvaselect{} mean ($\budget=8$) & 0.7698 & 0.7725 & 0.7758 \\
\midrule
Cycle $\Delta$ & \shortstack{$-0.0002$\\$[-0.0011,+0.0006]$} & \shortstack{$+0.0005$\\$[-0.0003,+0.0013]$} & \shortstack{$\mathbf{+0.0009}$\\$\mathbf{[+0.0003,+0.0015]}$} \\
Motion $\Delta$ & \shortstack{$+0.0006$\\$[-0.0003,+0.0016]$} & \shortstack{$-0.0006$\\$[-0.0014,+0.0001]$} & \shortstack{$-0.0004$\\$[-0.0009,+0.0001]$} \\
Consistency $\Delta$ & \shortstack{$\mathbf{+0.0016}$\\$\mathbf{[+0.0007,+0.0026]}$} & \shortstack{$\mathbf{+0.0006}$\\$\mathbf{[+0.0001,+0.0011]}$} & \shortstack{$+0.0000$\\$[-0.0004,+0.0004]$} \\
I2V $\Delta$ & \shortstack{$\mathbf{+0.0012}$\\$\mathbf{[+0.0003,+0.0022]}$} & \shortstack{$+0.0003$\\$[-0.0002,+0.0009]$} & \shortstack{$+0.0004$\\$[-0.0001,+0.0010]$} \\
\bottomrule
\end{tabular}
\end{table}
}

\newcommand{\paimainresulttable}{%
\begin{table}[H]
\centering
\setlength{\abovecaptionskip}{3pt}
\setlength{\belowcaptionskip}{2pt}
\caption{\textbf{PAI \cvaselect{} across generators ($\budget=8$).}  Official 8D
means over the same $174$ scenes and candidate identities.  Components:
Table~\ref{tab:pai-full}.}
\label{tab:pai}
\footnotesize
\setlength{\tabcolsep}{4.2pt}
\renewcommand{\arraystretch}{1.08}
\begin{tabular*}{\linewidth}{@{\extracolsep{\fill}}lccccc@{}}
\toprule
Model & \shortstack{no-TTS\\($\budget=1$)} & Proprio & VideoReward & \shortstack{\cvaselect{}\\w/o I2V} & \textbf{\cvaselect{}} \\
\midrule
Wan2.2-5B & 0.7594 & 0.7630 & 0.7639 & 0.7685 & \textcolor{cPick!85!black}{\textbf{0.7698}} \\
ABot-14B & 0.7689 & 0.7704 & 0.7686 & 0.7721 & \textcolor{cPick!85!black}{\textbf{0.7725}} \\
Cosmos3-Nano & 0.7730 & 0.7756 & 0.7731 & 0.7754 & \textcolor{cPick!85!black}{\textbf{0.7758}} \\
\bottomrule
\end{tabular*}
\end{table}

The common \cvaselect{} readout improves the no-TTS point estimate on all three
generators.  Its three frozen non-motion quality signals constitute G and the
oriented motion statistic constitutes M, so this grouping changes neither a
pick nor an official score.  Earlier historical and Top-3 variants remain
controls in Table~\ref{tab:pai-full}; they are not substituted into this row.
}

\newcommand{\robotwindiagnostictable}{%
\begin{table}[H]
\centering
\caption{\textbf{RoboTwin task-disjoint diagnostics ($\budget=8$).}
Official7 over $112$ scenes/$14$ tasks; task-bootstrap CIs; oracle is
evaluation-only.  Slow $-$ RM: $+0.000445$ $[-0.000209,+0.001371]$.}
\label{tab:robotwin-rm}
\footnotesize
\setlength{\tabcolsep}{2pt}
\begin{tabular}{@{}>{\raggedright\arraybackslash}p{0.23\linewidth}cccccc@{}}
\toprule
Method & $\budget$ & Off.7 & $\Delta$ zero & 95\% CI & Regret & W/T/L \\
\midrule
Baseline (no-TTS) & $1$ & $0.653234$ & $-0.001080$ & -- & $0.024388$ & -- \\
Native (zero memory) & $8$ & $0.654314$ & $0$ & -- & $0.023308$ & ref. \\
Pairwise RM$^\dagger$ & $8$ & $0.660807$ & $+0.006493$ & $[+0.003252,+0.010368]$ & $0.016815$ & $21/91/0$ \\
PM (held-out residual)$^\ddagger$ & $8$ & $\mathbf{0.661251}$ & $\mathbf{+0.006937}$ & $[+0.003430,+0.011082]$ & $\mathbf{0.016371}$ & $21/91/0$ \\
PM $+$ pretrain & -- & -- & -- & not measured & -- & -- \\
Oracle (eval. only) & $8$ & $0.677622$ & $+0.023308$ & -- & $0$ & -- \\
\bottomrule
\end{tabular}%
\vspace{1mm}
\parbox{0.97\linewidth}{\footnotesize $^\dagger$Post-outcome control; relocation
outcomes excluded from fit/picks.  $^\ddagger$Earlier outcome-blind diagnostic,
superseded by the fresh-episode audit.  Per-task:
Table~\ref{tab:robotwin-per-task}.}
\pmrobotvariants
\end{table}
}

\phantomsection\label{sec:ablation}

\newcommand{\physiqmemoryablationtable}{%
\begin{table*}[p]
\centering
\caption{\textbf{Physics-IQ memory ablation ($\budget=4$).}  Same $192$
scenes/$64$ families as Table~\ref{tab:crossbase}; LOFO excludes the query
family.  $^\ddagger$Privileged same-stem outcome.}
\label{tab:physiq-memory-ablation}
\footnotesize
\setlength{\tabcolsep}{5.2pt}
\renewcommand{\arraystretch}{1.08}
\begin{tabular}{@{}llcccc@{}}
\toprule
Variant & Information & Wan2.2-5B & ABot-14B & Cosmos-Nano & Avg. \\
\midrule
Baseline (no-TTS; $\budget=1$) & None & $25.07$ & $40.52$ & $27.72$ & $31.10$ \\
Low motion (zero memory) & None & $30.33$ & $43.69$ & $35.10$ & $36.38$ \\
\midrule
Held-out retrieval control & LOFO outcomes & $29.21$ & $41.93$ & $32.82$ & $34.65$ \\
G: train-global score & LOFO outcomes & $30.59$ & $44.13$ & $34.94$ & $36.55$ \\
N$+$M (without G) & LOFO outcomes & $29.14$ & $42.61$ & $34.58$ & $35.44$ \\
G$+$N (without M) & LOFO outcomes & $29.86$ & $43.20$ & $34.31$ & $35.79$ \\
\textbf{\cvaselect{} (selected)} & LOFO outcomes & $\mathbf{30.81}$ & $\mathbf{44.01}$ & $\mathbf{34.98}$ & $\mathbf{36.60}$ \\
\midrule
Paired-future bound$^\ddagger$ & Same-stem outcome & $\mathbf{32.19}$ & $\mathbf{45.67}$ & $\mathbf{36.83}$ & $\mathbf{38.23}$ \\
\bottomrule
\end{tabular}
\pmphysicsvariants
\end{table*}
}

\paragraph{Adaptive depth improves allocation, but not enough.}
Observable scores next choose stopping depth.  The best adaptive rules beat
random allocation on all three generators but not uniform sampling at the same
mean depth (Appendix Table~\ref{tab:entry-fee}).

\paragraph{A trajectory-internal action reaches the same boundary.}
On $179$ VideoPhy2-hard prompts, P\&P and uniform FlowMatch share the initial
latent and use $60$ denoiser evaluations.  P\&P lowers Joint success in all
three fresh-seed replicas, so the richer action still has no positive value
(Appendix Table~\ref{tab:trajectory-pnp}).

\subsection{A complete selector-component ablation}
We close the experiments by asking which observable components matter.  Every
non-empty subset reranks the same Physics-IQ pools under one information bound.
\input{tab_physics_ablation_main.tex}

Adding G or M changes the average by $+0.77$ or $+0.48$ IQ, whereas N changes
it by $-0.76$.  \cvaselect{} is only $+0.05$ over G alone, so the frozen-row reanalysis
motivates the common score without establishing synergy.

\section{Limitations}\label{sec:limitations}
The audit covers fixed pools, adaptive allocation, and one matched-NFE P\&P
intervention, not every learned policy.  Its failures localize tested actions;
they do not rule out video TTS.

The positive controls are also narrow: paired futures are privileged, PAI's
\cvaselect{} grouping is an exact renaming of frozen signals rather than a new
confirmation, the OpenS2V combination is post-hoc and negative, RoboTwin fails
fresh tests, and the PRM800K pass is sparse.  DiT-forward counts omit latency,
energy, memory traffic, and training cost, so each deployment must remeasure
opportunity and alignment.

All seven $B=4$ picks predate IQ, but \cvaselect{} was chosen after comparing them.
Without its family-level interval or $B=16$ picks, confirmation requires a
fresh sealed population.

\section{Conclusion}\label{sec:conclusion}
Across the studied video world models, more sampling creates oracle headroom
without reliably improving deployment: verifiers leave value unrecovered,
adaptive rules do not repay uniform compute, and matched-NFE P\&P remains
negative.  CVA nevertheless certifies the sparse PRM800K pass when state and
action align.

Proposer, verifier, and allocation policy therefore form one decision system:
scale compute only when an observable state supports an action whose gain
clears total cost.  Headroom is a possibility, not yet a return.

\newcommand{\reproducibilityappendix}{%
\section{Reproducibility details}\label{app:reproducibility}
This section traces the headline comparisons to their frozen candidate sets,
pick files, uncertainty procedures, and result hashes.  It is organized by
experiment rather than by implementation module so that each paper claim can be
followed from protocol to reported number.
\begin{figure}[t]
\centering
\includegraphics[width=\linewidth]{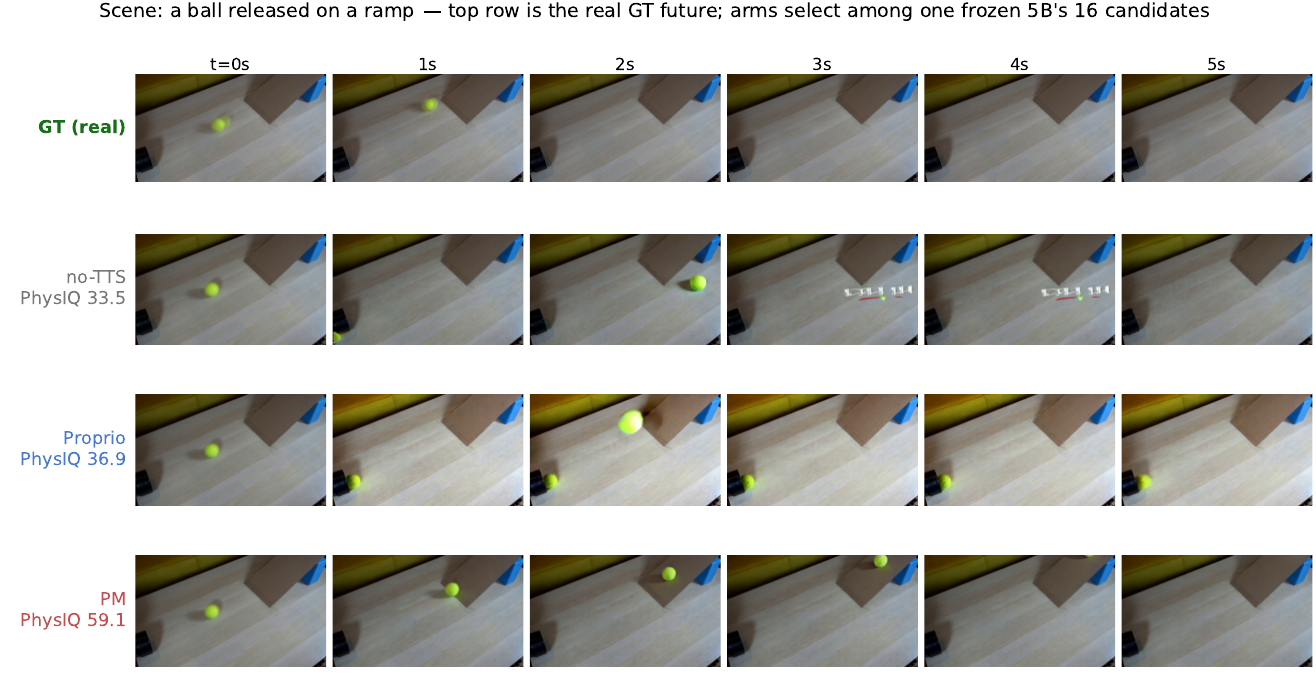}
\caption{\textbf{Qualitative selector comparison against the real future.}
\emph{Top:} the real Physics-IQ future; below, the same frozen
Wan2.2-TI2V-$5$B and conditioning frame.  The ball should leave the ramp:
no-TTS produces an overlay (IQ $33.5$), Proprio keeps it bouncing ($36.9$), and
the historical held-out retrieval control selects the correct departure
($59.1$).
This existing-$\budget=16$ example is representative, not an end-to-end
$\budget=4$ result or a typical-magnitude gain.}
\label{fig:qual}
\end{figure}
\paragraph{Broader positioning.}
Reasoning-time search includes repeated sampling, rank aggregation, tree search,
and world-model planning \citep{monkeys2024,llmblender2023,treeofthoughts2023,
rap2023,muzero2020,planet2019,pets2018}.  Video generation and evaluation span
Wan, VideoPhy, learned rewards, physics-aware steering, and sketch verification
\citep{wan2025,videophy2024,imagereward2023,videoscore2024,visionreward2024,
liu2025improving,wmreward2026,sketchverify2025}.  Persistent-state alternatives
include test-time adaptation, retrieval, episodic memory, and self-training
\citep{ttt2020,tent2021,cotta2022,rotta2023,t3a2021,knndiff2022,rdm2022,
nec2017,reflexion2023,expel2024,voyager2023,star2022,rest2023}.  Baseline-
relative decision rules are related to A-learning, residual control, safe policy
improvement, and reward mixtures \citep{alearning2014,residualrl2019,hcpi2015,
spibb2019,rewardedsoups2023}; \cva{} differs by auditing a frozen finite-pool
compute intervention rather than updating a policy or model.

The Physics-IQ comparisons use deterministic candidate seeds.  A result at
$\budget=4$ consumes candidates $c_{00}$--$c_{03}$; a result at $\budget=16$
consumes $c_{00}$--$c_{15}$.  Every consumed candidate is scored by the selector.
The held-out retrieval-control picks are materialized by
\texttt{eval/table1\_flow\_local\_template.py}: the 198-entry take-2 bank and
retrieval gate are hash-bound, picks are frozen without official IQ, and take-1
IQ is opened only in the analyze stage.  The complete G/N/M reanalysis uses the
seven already-frozen $B=4$ rows and is bound to
\texttt{eval/artifacts/physiq\_full\_factorial\_ablation\_20260903/}
\texttt{FULL\_FACTORIAL\_RESULT.json} (SHA256
\texttt{d2908103...0005b}).  It records \cvaselect{} as descriptive and explicitly
marks family-level uncertainty as unavailable; no $B=16$ \cvaselect{} result is
materialized.
The correlation-to-headroom test is materialized by
\texttt{eval/verifier\_law\_heldout.py}.  It fixes five deterministic folds over
the $66$ event families and evaluates the parameter-free identity on each held-out
fold; its slope, intercept, $R^2$, and MAE gates are declared in the script before
the result is computed.  Its uncertainty uses $5{,}000$ deterministic bootstrap
draws that resample event families within the frozen folds and retain all
world-model--verifier--budget points for each sampled family.  The motion-strata result is materialized by
\texttt{eval/physiq\_recurrent\_pm\_motion\_strata.py}: the median split is over
ground-truth event-family motion and is independent of method picks, and all CIs
bootstrap the $33$ held-out families rather than individual views.
The post-hoc disjoint-family calibration audit is materialized by
\texttt{eval/verifier\_law\_calibration\_split.py}: a SHA256 rule assigns
$27/39$ event families to calibration/evaluation; all families had entered the
earlier full-law analysis, while evaluation labels do not estimate $\rho$ in
this check.  It gives MAE $0.0697$ and Pearson $0.802$, while
leave-one-world-model-out MAE is $0.146$,
and the result SHA256 is \texttt{1662c6c9...259d0}.  The fixed-compute frontier
is materialized by \texttt{eval/verifier\_compute\_frontier.py} on the
$192$-scene/$64$-family complete-$16$ intersection.  It reports strict DiT NFE
and separately scoped shared-pipeline GPU-seconds with $5{,}000$ family
bootstrap draws; result SHA256 is \texttt{956e1f0b...d9f51}.
The headline width deltas and PAI comparisons are materialized by
\texttt{eval/paper\_paired\_uncertainty.py}.  It first reproduces every displayed
mean from frozen candidate identities, then uses $50{,}000$ paired event-family
draws for the $192$-scene Physics-IQ complete-$16$ intersection and paired scene
draws for all $174$ PAI scenarios; result SHA256 is
\texttt{06f894ee...b31f4}.
The PAI result is materialized by \texttt{eval/pai\_table2\_n8.py}; its frozen
artifact records $\budget=8$, $174$ scenes, all picks, input hashes, and complete
official eight-dimensional scores.  Its label-free table picks precede official
scores.  Equation~\eqref{eq:pai} proves that the displayed \cvaselect{} row is exactly
the original Native pick vector under a two-term grouping.  The separate
historical diagnostic uses preregistered five-fold cross-fitting over the same
$174$ PAI scenes.  In each fold, only the other four
folds' PAI seven-dimensional outcomes may write memory; held-out picks freeze
before their labels are opened, and no Physics-IQ inputs are loaded.  The
official eighth-dimension audit reuses exact candidate-identity scores for
$347/349$ required candidates and evaluates only the two remaining frozen picks;
\texttt{eval/pai\_fulln8\_memory\_crossfit.py} then verifies that the zero-memory control exactly
reproduces its Table~\ref{tab:pai} row before computing the paired historical
comparison.  The
developmental selector repair is materialized by
\texttt{eval/pai\_pairwise\_pm\_crossfit.py}: all $1{,}392$ candidate videos
receive label-free temporal descriptors; nested FIT/TUNE/CERT models freeze four
diagnostic pick sets before outer outcomes; and only the preregistered
\texttt{pairwise\_rich} arm is eligible as primary.  Its same-benchmark
representation amendment is explicitly not treated as independent confirmation.
For metric-matched auditing, frozen repaired policies reuse exact candidate-
identity eighth-dimension scores and evaluate only $20$ missing candidates;
the zero-memory control must reproduce the complete Table~\ref{tab:pai} value before any repaired
8D comparison is accepted.
The task-disjoint specificity campaign freezes DEQM and Candidate-DEQM picks
before new-task outcomes and compares them with matched shuffled-history
controls.  Their raw gains ($+0.006485$ and $+0.006211$) do not pass memory-
specificity gates.  The subsequent relocation study froze $112$ RoboTwin
scenes without outcomes, but the provenance record
\texttt{eval/PROVENANCE\_RESOLUTION\_ROBOTWIN\_V3\_V4\_20260716.md} marks it as
superseded diagnostic evidence because an authoritative follow-up was selected
and then stopped before execution.  The post-outcome ordinary pairwise RM
control is materialized by \texttt{eval/robotwin\_pairwise\_rm\_control.py}; it
uses no relocation labels for fit or picks and is explicitly non-confirmatory.
The PAI result uses exactly six changed interventions for inference; the
$174$-scene mean and five ID folds are descriptive only.  Two pre-metric plumbing
amendments bind the intended four-arm schema (including the diagnostic ungated
proposer) and strictly validate the complete $174\times16$ official cache before
extracting the frozen $c_{00}$--$c_{07}$ deployment subset.
The active model-forward and timing ledger is materialized by
\texttt{eval/active\_compute\_ledger.py}; it hashes all eight generation and
cycle logs and reports external reward-model calls separately from DiT NFE.
The fixed task-local fresh-episode confirmation uses episodes $0$--$3$ as
history and previously unscored episodes $8$--$11$ for evaluation on the same
$14$ tasks.  It
freezes all six arms and $256$ no-fixed-point wrong-task assignments before
creating the evaluation official7 scores.  The authoritative analyzer is
hash-frozen at
SHA256 \texttt{5475401f...4366525}; its result and the runner result are byte-
identical (SHA256 \texttt{ce1d6c41...7c043f}).  Task bootstrap, rather than
individual scenes, is the inference unit.
The untouched set-transport experiment freezes $48$ evaluation picks before official7;
its independently audited result SHA256 is
\texttt{c84f3f77...21d9c2d}.  The event-transition experiment likewise freezes
$56$ QUERY picks and both wrong-family and outcome-shuffle nulls before official7;
its independently audited result SHA256 is
\texttt{6331db5e...18fcb}.  Both analyses pass preregistered no-TTS and
Native-to-oracle-headroom validity gates and bootstrap tasks, not scenes.
The replacement episode-29 success-transition confirmation freezes all six
observed arms and 256 task derangements before materializing source futures or
official7.  Its freezer additionally requires zero SHA256 overlap against both
opened episode-28 candidate pools and source futures.  A pre-outcome extractor
ABI amendment changes only episode validation and is included in the 27-file
code manifest.  Frozen-pick, result, and independent-audit SHA256 values are
\texttt{06146daf...dd3fc}, \texttt{a0ac3527...7c0e0}, and
\texttt{813dc137...537e}, respectively.
The Memory Value Law implementation is materialized by
\texttt{eval/memory\_value\_law.py}; its final adapter reads only frozen picks and
subsequently opened official7 outcomes.  The $5{,}000$-draw task-cluster
artifact reconstructs both raw PM--RM deltas exactly and has SHA256
\texttt{f3d482c9...3862bf8a}.
}

\clearpage
\section*{AI Use Statement}
Generative AI tools were used as research assistants for exploring method
variants, implementing extraction and analysis code, automating experiments,
interpreting measured results, and drafting and editing this manuscript.  No AI
output was used as evidence: every reported number is produced by executable
code under a frozen protocol and bound to result artifacts.  The authors
verified all claims, take full responsibility for the content, and confirm that
no AI system is listed as an author.

\clearpage
\bibliographystyle{iclr2027_conference}
\bibliography{references}

\clearpage
\appendix
The appendix follows the same logic as the main audit.  It first exposes the
complete result tables, then expands the gate-localization and memory analyses,
and finally records provenance, developmental branches, and accounting checks.
\paimainresulttable
\begin{table}[H]
\centering
\caption{\textbf{Physics-IQ scaling.}  Complete $16$-candidate coverage;
$192$ scenes/$64$ families.  Each column generates and scores exactly
$\budget$ candidates; oracle is evaluation-only.}
\label{tab:oracle}
\footnotesize
\renewcommand{\arraystretch}{1.08}
\begin{tabular}{lcc}
\toprule
Selector & $\budget=4$ & $\budget=16$ \\
\midrule
Baseline (no-TTS; $\budget=1$) & \multicolumn{2}{c}{$25.07$} \\
Proprio (flow) & $25.30$ & $24.55$ \\
VideoReward & $25.76$ & $25.16$ \\
Native (cycle) & $27.53$ & $27.79$ \\
ABot-14B verifier & $26.59$ & $28.23$ \\
Low motion & $30.33$ & $31.90$ \\
\midrule
True-IQ oracle & $35.09$ & $44.32$ \\
\bottomrule
\end{tabular}
\pmwidthvariants
\end{table}

\section{Mechanism-level evidence for the audit}\label{app:cva-mechanism}
The main text reports where each population stops.  Here we expose the
mechanism-level comparisons behind those verdicts, beginning with the
cross-benchmark gate matrix and then the constructive reasoning case.
\input{tab_cross_benchmark_main.tex}

Table~\ref{tab:two-gates} records where each frozen evaluation stops.  The
diffuse attempt fails opportunity, MMLU-Pro fails action value, video stopping
reaches action value but not fee clearance, and PRM anchor--explorer clears all
four gates with sparse $7/128$ support.  This matrix is retrospective synthesis,
not a pooled benchmark score or an additional confirmatory experiment.

\input{tab_state_action_main.tex}
\input{tab_anchor_explore.tex}
\input{tab_anchor_ablation.tex}

\section{Does adaptive stopping repay its compute?}
\input{tab_entry_fee.tex}
Every observable rule is compared with uniform evaluated at each rule's
\emph{own} mean depth, so the fee is charged per rule.  Thus reversing five of the
six statistics beats random on each generator, yet no rule clears uniform.
The signal--remaining-headroom Spearman range runs from $-0.165$ to $+0.901$
across the eighteen statistic--pool pairs (sealed preregistration
\texttt{76a721b3}).  We measure that curvature rather than assume it; the reported
fee is measured, not signed by a theorem.  Bonferroni over all $132$
rule--budget--generator comparisons leaves no win for $\budget\ge4$.
Of these, $261$ of $7504$ resolve negative and $3$ resolve positive under
Bonferroni (largest $+0.183$), the remaining $96\%$ inside noise.  At the
$\budget=2$ boundary, three rules survive uncorrected, two
survive the within-cell correction and one survives the global one.  The raw
sign count says $43\%$ of second differences are positive on the deep pool
(sealed preregistration \texttt{5b4d5d75}).

\section{Compute-matched trajectory intervention}\label{app:trajectory-pnp}
The fixed-pool results do not determine whether intervention inside a sampling
trajectory can help.  We therefore translate the uncertainty-masked P\&P loop
of Self-Refining Video Sampling \citep{selfrefine2026} to Wan2.2-TI2V-5B
FlowMatch.  Uniform uses $60$ scheduler levels; P\&P uses $40$ base levels plus
$20$ fixed inner-loop predictions.  Both arms use the same prompt seed and
initial latent, exactly $60$ denoiser evaluations and $120$ CFG transformer
forwards, and no target feedback.  All $358$ videos are hash-bound before the
frozen VideoPhy2 SA/PC evaluator is opened.  Joint denotes
$\mathbf{1}[\mathrm{SA}\ge4]\mathbf{1}[\mathrm{PC}\ge4]$.

\begin{table}[H]
\centering
\caption{\textbf{Fresh VideoPhy2 P\&P audit.}  Each run contains all $179$
hard prompts; effects are paired percentage-point differences with $50{,}000$
prompt-bootstrap draws.  Replicas change only the prospectively frozen seed
salt.  $P_+$ is the bootstrap probability of a positive effect.}
\label{tab:trajectory-pnp}
\footnotesize
\setlength{\tabcolsep}{3.5pt}
\begin{tabular}{@{}lccccc@{}}
\toprule
Run & Uniform Joint & P\&P Joint & $\Delta$ Joint $[95\%\ \mathrm{CI}]$ & $P_+$ & sign $p$ \\
\midrule
Primary & $5.03\%$ & $3.35\%$ & $-1.68\ [-4.47,+1.12]$ & $.083$ & $.938$ \\
Replica 1 & $6.70\%$ & $4.47\%$ & $-2.23\ [-5.59,+1.12]$ & $.073$ & $.945$ \\
Replica 2 & $6.15\%$ & $4.47\%$ & $-1.68\ [-4.47,+0.56]$ & $.047$ & $.969$ \\
Replica 3 & $5.59\%$ & $5.03\%$ & $-0.56\ [-4.47,+2.79]$ & $.324$ & $.726$ \\
\bottomrule
\end{tabular}
\end{table}

The preregistered success rule requires every primary gate to pass; the primary
run fails action-value, uncertainty, sign, PC, SA-noninferiority, and groupwise
gates.  The robustness rule additionally requires at least two nonnegative
replicas and a positive mean.  Instead, $0/3$ are nonnegative and their mean is
$-1.49$ pp.  Compute and basic safety checks do pass: runtime ratios are
$0.998$--$1.000$, motion ratios are $1.327$--$1.390$, sharpness ratios are
$0.964$--$1.057$, and black-frame deltas are zero.  The primary result SHA-256
is \texttt{bb57a1de...0a65e0d}; the three replica results are
\texttt{0bdcfdcf...55050f6}, \texttt{bfacd0ee...bd7e1}, and
\texttt{3c263e99...4a4740a}.

\section{How the evidence changed during development}
Because several exploratory branches preceded the frozen analyses, this section
separates retained evidence from interpretations that were later narrowed or
withdrawn.  The purpose is provenance, not an additional aggregate result.
The deep-pool effect is estimable, but it still shows no gain; we therefore
withdraw its first interpretation.  The accounting identity is
$\mathrm{Overall} \equiv \mathrm{opportunity}+\mathrm{alignment}$.
Historical transfer diagnostics record clean-pair $0.969$, MQ-to-TA $-0.387$,
convenience-sample $0.318$, sizing floor $26.6$, and sizing constant $1.423$.
They establish a video boundary, not a universal law; the earlier unification
was too strong.  Verifier audits record the pooled/within-pool flip $-0.139$,
pooled-validation cost $+0.494$, orientation-only share $71.5\%$, harmful
selector loss $-0.398$, and prescription cost $0.0823$.  The refuted
pool-limited prediction is target-specific.  The variance extension resolves
$11/15$ pairs with exact covariance residual $1.8\times10^{-16}$; the fixed
effect has a local VQ win $+0.063$ but shuffled-ridge control $-0.150$.

For reasoning, anchor--explorer passes its constructive audit, with greedy
no-TTS at $73.44\%$; this does not establish a positive adaptive video
selector.  The attempted transfer to Qwen3.5 is invalid.  The MMLU-Pro evidence
does confirm a gap between state confidence and action value, but the one-point
compound prediction is not confirmed, and the official-CoT smoke is invalid at
$78.57\%$.  Video fee localization recovers only $41.6$--$69.0\%$.

\paragraph{Frozen verdict index.}
For exact artifact lookup, the retained statuses are
\texttt{estimable-but-still-no-gain},
\texttt{ANCHOR\_EXPLORE\_CONSTRUCTIVE\_PASS},
\texttt{INVALID\_QWEN35\_ANCHOR\_TRANSFER},
\texttt{MMLUPRO\_STATE\_ACTION\_GAP\_CONFIRMED},
\texttt{INVALID\_ANCHOR\_DIFFUSE\_REPLICATION}, and
\texttt{GATE\_LOCALIZATION\_MATRIX\_CLOSED}.  Greedy is not necessary for the
reasoning positive control, which is not a positive adaptive video selector.
The video development stops are
\texttt{KILL\_DIRECT\_SET\_ACTION\_VALUE\_BEFORE\_FRESH} and
\texttt{CURRENT\_ACTION\_VALUE\_INFORMATION\_CLASS\_CLOSED}; Unified visual
Top-3 (retrospective) remains a scoped PAI control rather than a prospective
method claim.

\paragraph{What the cross-regime comparison does and does not show.}
The cross-regime state/action matrix is a fixed retrospective synthesis: it
localizes which CVA gate failed but is not an additional confirmatory result.
Estimability is a preflight condition, not an empirical result.  This post-hoc
state--action localization cannot broaden a population claim; when no
alternative action exists, the trap cannot arise.  It creates no entry in the
paper's evidence base beyond the frozen findings reported above.

\paragraph{Why the diffuse replication does not support generality.}
The independently audited diffuse attempt has only $+5.86$ points of oracle headroom,
$12/256$ positive contributions, and a top-$10\%$-trimmed effect of $-0.42$
points.  The replication is invalid and supports neither diffuse nor
cross-source generality.

\paragraph{Why VideoPhy2 stops at action value.}
The direct structured set predictor fails its action-value gate before any
fresh evaluation: its advantage is $+0.27$ points with $p=.572$, while the
target-oracle routing advantage is $+3.07$ points.  This does not justify a
fresh window.  Across the broader
representation ladder, cross-source attention reaches $+1.67$ points
($p=.160$), cross-target transfer has $p=.385$, and explicit source-window
action supervision has $p=.825$.  Together, these results close the tested
action-value information class and do not justify fresh video generation.
A candidate-specific $8/30$-block partial-DiT preview reaches the same boundary:
after a fixed $c00$--$c07$ fit, its one-shot $c08$--$c15$ Q delta against the
training-selected VideoReward control is $-0.123$ (CI $[-0.201,-0.045]$;
full-refit null $p=.568$).  It is killed before fresh generation.

\section{PM implementation details and evidence boundary}\label{app:pm-details}
The main text treats PM as an audited route.  This section makes that boundary
operational by separating legal information, matched controls, and privileged
references before detailing each benchmark-specific adapter.
\physiqmemoryablationtable
\begin{table*}[t]
\centering
\caption{\textbf{Memory-independence audit.}  Attribution requires valid
information boundaries and matched-history controls.  PIQ: Physics-IQ; RT:
RoboTwin; $95\%$ CIs.}
\label{tab:memory-independence}
\footnotesize
\setlength{\tabcolsep}{2.5pt}
\renewcommand{\arraystretch}{1.02}
\begin{tabular}{@{}
>{\raggedright\arraybackslash}p{0.12\linewidth}
>{\raggedright\arraybackslash}p{0.31\linewidth}
>{\raggedright\arraybackslash}p{0.34\linewidth}
>{\raggedright\arraybackslash}p{0.15\linewidth}@{}}
\toprule
Setting & Effect & Control & Verdict \\
\midrule
PIQ held-out & Across WMs $+0.44\ [-0.77,1.57]$ & LOFO; train-global $+1.87$ vs retrieval & Transfer killed \\
PAI five-fold & \shortstack[l]{$-0.000197$\\$[-0.000493,+0.000021]$} & Train-only writes; frozen outer folds & Memory killed \\
RT DEQM & $+0.006485$; $9$W/$0$L & History shuffle & Specificity fails \\
RT Candidate-DEQM & $+0.006211$; $6$W/$0$L & Fixed-memory shuffle & Specificity fails \\
RT fresh local & \shortstack[l]{$+0.001514$\\$[-0.002031,+0.005258]$} & Eps. 8--11; 256 wrong-task nulls & Residual killed \\
RT transition (e29) & \shortstack[l]{Region vs RM $+0.058848$\\$[0.016109,0.103457]$\\Off.7 $-0.007112$\\$[-0.011187,-0.003448]$} & No candidate/source alias; wrong-task/shuffle pass; global CI crosses $0$ & Region passes; VRM killed \\
RT extended (dev.) & \shortstack[l]{Task--global $-0.003324$\\$[-0.008816,+0.002076]$\\$3$W/$5$T/$6$L} & Eps. 12--15$\to$16--19; wrong-task/shuffle pass & Task ID killed \\
RT reliability (dev.) & \shortstack[l]{Off.7 vs global $+0.002655$\\$[0.000522,0.005227]$\\$4$W/$10$T/$0$L} & Native LOEO; wrong-task CI $>0$; shuffle $p=.003891$ & Model choice passes; replication killed \\
RT stopping family & Mean $B$: initial/predictive/barrier/reliability $5.964/8.000/5.571/5.607$; content-global $5.429$; shrunk task/global $5.964/6.321$ & Split calibration; global/wrong-task/256 shuffles; write-stage audit & Content risk helps; shrunk-task CIs cross $0$; nearest-neighbor retrieval fails \\
RT cross-WM & \shortstack[l]{5B PM--RM $0\ [0,0]$\\14B $-0.000705$\\$[-0.002508,+0.000639]$} & Frozen eps. 16--19; matched RM & Slow PM killed \\
RT STOM & $-0.001046\ [-0.003138,0]$; $0/46/2$ & 48 sealed scenes; wrong-task/outcome shuffles & Transport killed \\
RT event transition & $0\ [0,0]$; $0/56$ changed & 56 sealed queries; global/shuffle & Utility killed \\
\bottomrule
\end{tabular}
\pmcrossvariants
\end{table*}

\paragraph{Proof and information boundary for Proposition~\ref{prop:memory-value}.}
Add and subtract $\langle r^*\rangle_h$:
\begin{equation}
\langle r_\pi-r_{\pi_0}\rangle_h
=\langle r^*-r_{\pi_0}\rangle_h
-\langle r^*-r_\pi\rangle_h.
\end{equation}
Taking $\pi_0$ to be
$a^*_{\rm fix}=\arg\max_a\langle r_a\rangle_h$ gives $G$; nonnegativity follows
from $\langle\max_a r_a\rangle_h\geq\max_a\langle r_a\rangle_h$.
Conditionally applying Proposition~\ref{prop:verifier-quality} at each $x$
gives $r_a(x)=\rho_a(x)$ and
therefore
\begin{equation}
V_{\rm mem}
=\langle\max_a\rho_a\rangle_h-\max_a\langle\rho_a\rangle_h
-\langle\max_a\rho_a-\rho_{\pi}\rangle_h.
\end{equation}
The headroom weights are necessary when candidate quality scales differ across
conditions.  For the information bound, let $P_h$ denote this headroom-weighted
distribution.  For each $(c,m)$, the advantage of the best conditional action
over the best action given $c$ is at most
$L\operatorname{TV}(P_h(R\mid c,m),P_h(R\mid c))$.  Taking expectations and
applying Pinsker's inequality followed by Jensen gives
$L\sqrt{I_h(R;M\mid C)/2}$.  Hence $R\perp M\mid C$ implies zero population
value over the Bayes-optimal history-blind policy.  These are population
identities, not a claim that the reliability-memory proposal has
passed empirically.

\paragraph{Decision value and measured action-space checks.}
For arbitrary action utilities $U_a$, Bayes-optimal memory has gross value
\begin{equation}
V^*_{M\mid C}=\mathbb E\!\left[\max_a\mathbb E[U_a\mid C,M]
-\max_a\mathbb E[U_a\mid C]\right]\geq0.
\label{eq:bayes-memory-value}
\end{equation}
Conditional Jensen proves nonnegativity; equality holds whenever one
$C$-measurable action is posterior-optimal almost surely.  Thus memory may
predict outcomes without changing the optimal action, and observation,
retrieval, or compute costs may make net value negative.

On label-sealed episode 29, mean random-to-oracle region headroom is $0.1386$;
normalized addressable value $0.4855$ minus retrieval regret $0.0609$ gives
$0.4246$ over Ordinary and reconstructs the raw $+0.0588$ region gain.  Its
simultaneous official7 loss makes this proxy-target value non-deployable.  In
the retrospective cross-model LOEO audit, history routes between frozen 5B and
14B Native actions.  Routing beats global by $+0.002655$ official7 (task CI
$[+0.000522,+0.005227]$), cyclic wrong-task routing (CI
$[+0.002501,+0.015403]$), and 256 task derangements ($p=0.003891$).  Its law is
$0.004024$ addressable value minus $0.001369$ regret.  However, all four
e20--23$\to$e24--27 temporal replications fail, including the strict
e16--19/e20--23/e24--27 train/validation/test chain; hence this is developmental,
not fresh confirmation.

\paragraph{Applying the law to frozen cross-model picks.}
We apply Eq.~\eqref{eq:memory-value-law} to the frozen cross-world-model picks,
including Native, Semantic PM, and the same-data ordinary RM in the local
envelope and using RM as $\pi_0$.  On 5B, Semantic and RM are identical, so
addressable value, regret, and memory value are all $0$.  On 14B, normalized
addressable value is $0.0679$, retrieval regret is $0.1093$, and memory value is
$-0.0413$ (14-task cluster CI $[-0.1435,0.0390]$).  Multiplying by mean headroom
$0.0171$ exactly reconstructs the raw PM--RM difference $-0.000705$.

Section~\ref{sec:pm} defines the verifier interface, causal memory state, and
slow/fast composition.  For a label-free candidate descriptor $x(z)$, the
reward-tier slow state fits prior within-scene outcome residuals,
\begin{equation}
\begin{aligned}
\beta_t=\argmin_\beta\;&\sum_{j<t}\sum_{a<b}w_{jab}\bigl[
(y_{ja}-y_{jb})-(u_v(z_{ja})-u_v(z_{jb}))\\[-1mm]
&\hspace{34mm}-(x(z_{ja})-x(z_{jb}))^\top\beta\bigr]^2
+\gamma\|\beta\|_2^2 .
\end{aligned}
\label{eq:pm-slow}
\end{equation}
This is a compact outcome-trained reranker, not a world-model update.  Its matched
ordinary pairwise-RM control is therefore essential.  In the final frozen
episodes 16--19 audit, the two rules are identical on 5B and Semantic Slow PM
is $-0.000705$ below ordinary RM on 14B (CI
$[-0.002508,+0.000639]$); thus both 95\% upper bounds exclude a $+0.001$ gain.

\paragraph{Physics-IQ: observed-outcome memory.}
For same-stem repeated setup, a previously observed take-2 real outcome $\hat T_i$
enters through
\begin{equation}
m_{\mathrm{PIQ},i}(z)=
Z[\operatorname{IoU}(M(z),M(\hat T_i))]
-Z[d_{\mathrm{app}}+d_{\mathrm{temp}}+d_{\mathrm{motion}}].
\label{eq:outcome-memory}
\end{equation}
Physics-IQ take-2 supplies $\hat T_i$ and take-1 remains the evaluation
realization.  Both the history key and query key are computed from the same
canonical $c_{00}$ conditioning file for each stem; the resulting $198/198$
match verifies key wiring, not retrieval generalization or pixel-identical
take-1/take-2 first frames.  This is the sole positive historical boundary,
but it consumes extra predictive information.
Family-held-out retrieval is worse than a train-global history control.

\begin{table}[htbp]
\centering
\caption{\textbf{Privileged same-stem Physics-IQ reference ($\budget=4$).}
Means and $95\%$ family-bootstrap CIs.  The paired take-2 future is an
appendix-only privileged diagnostic.}
\label{tab:piq-privileged-outcome}
\footnotesize
\setlength{\tabcolsep}{4pt}
\renewcommand{\arraystretch}{1.08}
\begin{tabular}{@{}lccc@{}}
\toprule
Selector & Wan2.2-5B & ABot-14B & Cosmos-Nano \\
\midrule
Native (no history) & \shortstack{$27.48$\\$[23.65,31.45]$} & \shortstack{$41.16$\\$[35.71,46.76]$} & \shortstack{$33.62$\\$[28.97,38.40]$} \\
PM (held-out) & \shortstack{$29.11$\\$[24.79,33.47]$} & \shortstack{$41.67$\\$[36.28,47.25]$} & \shortstack{$32.80$\\$[28.17,37.46]$} \\
PM $+$ pretrain (paired; priv.) & \shortstack{$32.16$\\$[27.80,36.62]$} & \shortstack{$45.27$\\$[39.58,51.09]$} & \shortstack{$36.87$\\$[31.92,41.99]$} \\
\textit{$\Delta$: pretrain $-$ held-out} & $+3.05$ & $+3.60$ & $+4.07$ \\
\bottomrule
\end{tabular}
\pmphysicsallvariants
\end{table}

The privileged reference exceeds Native by $+4.68/+4.12/+3.25$ IQ, with paired
$95\%$ CIs $[+2.94,+6.53]/[+2.38,+6.15]/[+1.82,+4.71]$.  A preregistered
Flow-plus-history fusion lowers the outcome-only reference by
$1.71/2.24/2.15$ IQ.  These numbers measure the value of an exactly paired real
future, not held-out memory generalization: family-held-out retrieval has a
cross-world-model delta of $+0.44$ with CI $[-0.77,1.57]$ and is worse than the
train-global outcome control by $-1.87$ IQ with CI $[-3.28,-0.64]$.

\begin{table}[htbp]
\centering
\caption{\textbf{Privileged Physics-IQ motion strata ($\budget=4$).}
$33$ calm/$33$ dynamic families ($99$ scenes each); $95\%$ CIs use $50{,}000$
family bootstraps.  Hist. uses paired take-2 and is non-headline;
$\Delta=\mathrm{Hist.}-\mathrm{Low}$ on dynamic families.}
\label{tab:motion-strata}
\footnotesize
\setlength{\tabcolsep}{2.0pt}
\renewcommand{\arraystretch}{1.08}
\begin{tabular}{@{}lccccccc@{}}
\toprule
& \multicolumn{2}{c}{Calm} & \multicolumn{4}{c}{Dynamic} & \\
\cmidrule(lr){2-3}\cmidrule(lr){4-7}
Model & Low & Hist. & no-TTS & Native & Low & Hist. & $\Delta$ [95\% CI] \\
\midrule
Wan2.2-5B & $31.54$ & $32.09$ & $25.97$ & $25.57$ & $28.93$ & $32.24$ & $+3.31\ [1.47,5.21]$ \\
ABot-14B & $46.93$ & $46.82$ & $40.80$ & $39.99$ & $39.86$ & $43.72$ & $+3.86\ [2.13,5.77]$ \\
Cosmos3-Nano & $33.23$ & $33.78$ & $32.48$ & $35.78$ & $36.87$ & $39.96$ & $+3.08\ [1.37,4.92]$ \\
\bottomrule
\end{tabular}
\pmphysicsallvariants
\end{table}

\paragraph{PAI \cvaselect{} readout.}
PAI-Bench-robot rewards perceptual consistency and conditioning faithfulness in
addition to plausible motion.  For its unified $\budget=8$ candidates, define
G as the three non-motion quality signals and M as its positively oriented
motion-adequacy statistic.  All terms are standardized within scene:
\begin{equation}
\underbrace{q_{\mathrm{PAI}}(z)}_{\text{\cvaselect}}
=\underbrace{Z[-q_{\mathrm{cyc}}(z)]
 +Z[q_{\mathrm{consist}}(z)]+Z[q_{\mathrm{i2v}}(z)]}_{G(z)}
 +\underbrace{Z[q_{\mathrm{motion}}(z)]}_{M(z)}.
\label{eq:pai}
\end{equation}
These cheap video/conditioning statistics use no external VLM or PAI label.
Equation~\eqref{eq:pai} is algebraically identical to the previously named
Native readout, so the unified name does not retrofit new numbers.  The
cross-base ablation removes one constituent signal at a time.  Historical
adapters use \cvaselect{} only as their explicit zero-memory null and remain separate
appendix controls.

\paragraph{PAI/RoboTwin: reward-tier memory.}
These benchmarks provide no model-free observed future.  Their legal analogue
is Eq.~\eqref{eq:pm-slow} applied to task-disjoint prior candidate outcomes with
matched outcome-shuffle or wrong-task controls.  Under that protocol, PAI
cross-fit, DEQM specificity, fresh task-local adaptation, untouched set
transport, and event-transition readout all fail their gates.

A preregistered episodes 8--10 $\rightarrow$ 28 diagnostic does improve
held-out transition-region score over ordinary RM by $+0.075982$ (CI
$[+0.028359,+0.129885]$).  It nevertheless fails the wrong-task,
global-history, strict-win, and official7-noninferiority gates, so its
task-specific retrieval claim is killed.  A subsequent integrity audit also
found that 72/112 candidate MP4 hashes were byte-identical to an earlier pool
whose official7 scoring had opened before these picks froze.  The E28 result is
therefore diagnostic rather than label-sealed fresh evidence.

Conformal Memory Stopping was then strengthened in three fixed stages.  Numeric
predictive conformal remains at B8.  Selection-aware Barrier CMS lowers mean
candidates from $5.964$ to $5.571$ (global $5.982$), is quality-noninferior to
B8, and beats wrong-task memory with a positive CI.  Its gain versus global is
nevertheless $+0.000732$ with CI $[-0.001756,+0.003533]$, below the shuffle
p97.5; its tail and oracle-hit gates also fail.  A disjoint reliability gate
further reduces the effect.  The post-outcome safe two-arm oracle saves $0.518$
candidates, and the safe-prefix oracle saves $2.964$.  These oracles establish
remaining capacity, but not a deployable claim.

Adding prefix content---verifier, latent, and temporal coverage---improves the
global stopper from $5.982$ to $5.429$ candidates, raises quality by
$+0.001746$, halves the violation rate from $16.1\%$ to $8.9\%$, and increases
oracle hits from 12 to 18.  However, the quality and compute-aware CIs narrowly
cross zero, calibration AUC rises only $+0.0045$, and the original two-episode
task-memory residual makes the content-aware arm worse.  A fixed repair writes
six leave-one-episode-out residuals per task, retains only sign-stable one-sided
safe surplus, and applies empirical-Bayes shrinkage.  This SS-EB task residual
reduces candidates from $6.321$ to $5.964$ and flips the compute-aware
development effect to $+0.001314$ (CI $[-0.000230,+0.003191]$), beating 256
task derangements ($p=.007782$).  Its CI versus wrong-task still crosses zero
($+0.000874$, CI $[-0.000753,+0.002456]$).

A subsequent audit, conducted before test outcomes are opened, rejects
three-neighbor content retrieval: its out-of-fold Brier score is $0.167102$
versus $0.164665$ for the task average.  Thus the
harmful residual is repaired at the point-estimate level, but neither the
representation nor task-memory attribution is confirmed; fresh episode 30 is
not authorized.

The replacement label-sealed episode-29 confirmation closes the loop on a
genuinely fresh pool.  It has 112 unique candidate hashes and 14 source hashes
with zero overlap against both opened episode-28 pools.  Success-region
alignment improves over Ordinary by $+0.058848$ (CI
$[+0.016109,+0.103457]$), beats wrong-task memory (CI
$[+0.006397,+0.096932]$), and beats 256 task derangements ($p=0.003891$), with
eight strict joint wins.  Yet it remains indistinguishable from global success
memory ($+0.000866$, CI $[-0.019748,+0.019357]$) and reduces official7 versus
Ordinary by $-0.007112$ (CI $[-0.011187,-0.003448]$).  The result therefore
confirms the success-region proxy while killing task-specific, quality-safe
VRM.

The final frozen-pick audit then uses all $14$ tasks, episodes 16--19, and $56$
scenes per world model.  On 5B, both Semantic Slow PM and ordinary RM improve
over Native by $+0.005805$, but their picks and means are exactly identical
(PM--RM $0$, CI $[0,0]$).  On 14B, PM--Native is $+0.001086$ (CI
$[-0.001728,+0.004211]$), whereas PM--RM is $-0.000705$ (CI
$[-0.002508,+0.000639]$).  A Joint-Null Cone variant is inconclusive on aligned
5B ($+0.000106$ vs Native, CI $[-0.000994,+0.001209]$); its positive 14B result
cannot rescue the failed aligned and cross-world-model gates.  The sealed
result SHA256 values are \texttt{fa7495f0...dcc33d} (matched-RM control) and
\texttt{5c2f0528...1813bd} (Joint-Null Cone).

\paragraph{Current-state capability routing.}
An e16--19 train/e20--23 validation chain tests six increasingly direct keys:
frame-0 GME; a 1370D generation-time robot/scene/image state; its TRAIN-only PCA
bottleneck; 107D explicit foreground geometry; paired 5B/14B candidate-0
responses; and an equal-compute quota ranking.  None passes its familywise gate.
Signed routing has nine oracle corrections across seven tasks but only
$0.001275$ addressable value, so a $+0.001$ target requires $78.4\%$ recovery.
At a fixed 16/56 5B quota, the oracle improves
task routing by $+0.001093$ (CI $[+0.000090,+0.002504]$) while retaining
$21.77\%$ compute-proxy savings, but pilot memory makes zero swaps.  A fixed
nearest-history audit explains the failure: same-position within-task rank
concordance is $0.607$ ($p=.0817$), and the best of seven state mappings has
residual Pearson $0.081$ versus familywise null p97.5 $0.306$ ($p=.704$).
Even the best direct mapping loses $0.001528$ to task-only ($p=.611$): it fixes
five crossover errors but breaks four correct actions.  Geometry fixes five and
breaks seven; each pilot view fixes three and breaks seven.  Reserved e24--27
features and outcomes remain unopened.  This closes identification for the
frozen signals tested here, but not for a future fresh early-denoise
uncertainty probe.

Table~\ref{tab:memory-independence} is the complete main-text claims ledger;
Appendix~\ref{sec:pai-developmental} reports the earlier diagnostics.

\robotwindiagnostictable

\paragraph{OpenS2V-Eval: the same interface, a different outcome.}
The OpenS2V reanalysis uses four matched candidate sources per task.  G is the
frozen same-task cross-seed component-mean utility; M is negative frame-
difference motion on the fresh video.  Their equal-weight standardized sum
selects without opening official metrics.  Because this combination was
specified after the source results existed, Appendix Table~\ref{tab:opens2v-pilot}
reports it as descriptive evidence, not confirmed transfer.

\section{Prospective pretrained global capability memory}
\label{app:pgcm}
The failed target-history adapters suggest that repeated target writes are not
the right source of value.  We therefore examine a stricter alternative: learn one
task-agnostic memory once on a public source distribution, freeze it, and
transfer it without target outcomes.  We preregister \emph{Pretrained Global
Capability Memory} (PGCM).  A one-step, no-decode probe from frozen 5B and 14B
models yields $27+27+27=81$ scalar features (5B, 14B, and their difference).
The original source-only ridge predicts Native 14B-minus-5B advantage, but a
pooled parametric predictor is global utility rather than memory.  We therefore
make it a matched control and define the memory as one shared public bank
\begin{equation}
\mathcal M_{\rm pub}=\{(k_i,v_i)\}_{i=1}^{192},\qquad
k_i=P_d\,\mathrm{Std}(z_i),\quad
v_i=Q_i^{14\mathrm B}-Q_i^{5\mathrm B},
\label{eq:pgcm-bank}
\end{equation}
where $z_i\in\mathbb R^{81}$ is the paired probe.  A query retrieves at most
one view per source family and uses a Gaussian-weighted mean of the nearest
stored values.  PCA dimension $d\in\{8,16,32,64,81\}$, distinct-family
neighbor count $K\in\{4,8,16,32\}$, and temperature
$\tau\in\{0.25,0.5,1,2\}$ are chosen by nested leave-one-source-family-out
prediction.  The final transform, bank, and 5B quota are then refit once on all
$64$ public families and frozen.  No task identifier, prompt text, target
history, target example, or target label enters this shared memory.

\begin{table}[htbp]
\centering
\caption{\textbf{PGCM attribution.}  Source arms share probes, labels, folds,
candidates, and quota.  Memory must beat ridge and both matched nulls.}
\label{tab:pgcm-pretrain}
\footnotesize
\setlength{\tabcolsep}{2.4pt}
\begin{tabular}{@{}
>{\raggedright\arraybackslash}p{0.21\linewidth}
>{\raggedright\arraybackslash}p{0.25\linewidth}
>{\raggedright\arraybackslash}p{0.16\linewidth}
>{\raggedright\arraybackslash}p{0.28\linewidth}@{}}
\toprule
Arm & Source object & Target labels & Role \\
\midrule
Source-majority & Mean advantage & None & Feature-free null \\
Global ridge & 81-D parametric fit & None & Utility control \\
Global memory & Shared key/value bank & None & Nonparametric memory route \\
Value/key shuffle & Deranged bank & None & Specificity nulls \\
Target-history ridge & Private fit & Used & Supervised upper comp. \\
\bottomrule
\end{tabular}
\pmpgcmvariants
\end{table}

On nested source OOF predictions, global memory must beat global ridge by at
least $+0.25$ IQ with a family-bootstrap CI lower bound above zero and exceed
the 97.5th percentile of $256$ family-value shuffles and $256$ family key/value
mismatches.  Otherwise the memory claim is killed even if ridge transfers.

\begin{table}[htbp]
\centering
\caption{\textbf{Outcome-free transfer ladder.}  One frozen memory throughout.
PAI outcomes are open; confirmation requires a new owner-held split.}
\label{tab:pgcm-protocol}
\footnotesize
\setlength{\tabcolsep}{2.4pt}
\begin{tabular}{@{}
>{\raggedright\arraybackslash}p{0.15\linewidth}
>{\raggedright\arraybackslash}p{0.25\linewidth}
>{\raggedright\arraybackslash}p{0.24\linewidth}
>{\raggedright\arraybackslash}p{0.26\linewidth}@{}}
\toprule
Tier & Data & Shift & Role \\
\midrule
Pretrain & Physics-IQ, $192$/64 & Natural events & Source fit \\
Near & Held-out PIQ families & New families & Public diagnostic \\
Mid & OpenS2V-Eval & New scenes/interface & Public diagnostic \\
Far-dev. & Private PAI, $174$ & Robot I2V/official8 & Retrospective \\
Far-confirm. & New robot, $\geq100$ & Owner-held assets & Claim required \\
\bottomrule
\end{tabular}
\pmpgcmvariants
\end{table}

PAI is selected before quality scoring because it changes event semantics
(natural interactions to robot manipulation), scene/camera composition,
evaluation target (IQ to official8 perceptual and conditioning quality), and
visibility (public videos to private condition images).  RoboTwin is public and
is therefore ineligible for the private-target claim.  ``Far OOD'' additionally
requires zero image/task overlap and two independent outcome-free audits: both
a frozen DINOv2 ViT-L/14 condition-image encoder (fixed 518 center crop, 1024-D
CLS token, frozen revision/weight hash) and the 81-D denoise probe must obtain
group-held-out domain AUC $\geq0.80$ and group-mean linear-MMD permutation
$p\leq0.01$.  Failure of either audit removes the far-OOD claim regardless of
quality.

The primary target controls are global ridge, source-majority routing,
equal-quota random routing, both memory shuffles, always-5B/14B, and an
equal-quota oracle.  Global memory must improve both ridge and source-majority
by at least $+0.0005$ official8 with positive paired uncertainty, pass the
matched nulls, remain noninferior to always-14B, increase oracle hits, and save
compute after charging both probes.  The full frozen gates are in
\texttt{eval/PREREG\_pretrained\_global\_capability\_memory\_20260720.md} and
its two amendments, including
\texttt{AMENDMENT\_pgcm\_shared\_memory\_ood\_ladder\_20260720.md}.

Real outcome-free PGCM probes were completed for all $192$ public source
scenes, but both source estimands fail their memory-specific gates.  In the
independently frozen Table-1-matched $B=4$ estimand, memory minus ridge is
$+0.00$ IQ (95\% CI $[0.00,0.00]$), with value-shuffle $p=0.9455$ and
key/value-mismatch $p=0.9728$.  In the separately frozen original $B=8$
transfer-source estimand, memory minus ridge is again $+0.00$ IQ (95\% CI
$[0.00,0.00]$), with value-shuffle $p=0.8911$ and key/value-mismatch
$p=0.8444$.

Both primary results use the frozen analyzer's summed squared Euclidean
distance in the selected PCA space; its code hash is bound in the analysis
input receipt.  Because the B4 source-memory attribution gate fails, no
claim-eligible Table-1 row is inserted.  A post-stop diagnostic using the
clipped 1/192 5B quota scores $41.56$ and differs from held-out 14B PM by
$-0.36$ IQ (95\% CI $[-1.80,1.06]$); it is not claim-eligible.  These outcomes
are public same-domain held-family-out evidence only.  Private PAI labels remain
open, and owner-held confirmation still requires a new private split of at
least $100$ scenarios.

\section{Memory-conditioned seed screening and velocity diagnostics}
\label{app:seed-screening}
The preceding experiments rank completed candidates.  This section asks whether
history can save generation compute earlier, while preserving natural-noise
sampling and guarding against off-manifold optimization.
\paragraph{Select natural noise; do not rewrite it.}
For deterministic sampling $z^{(b)}=F_\phi(x,\epsilon^{(b)})$, selecting a
completed candidate also selects its initial seed, but only after paying for the
rollout.  We preregister the compute-saving variant
\begin{equation}
\epsilon^{(1:M)}\!\stackrel{\mathrm{iid}}{\sim}\mathcal N(0,I),\quad
h_\tau^{(b)}=F_{\phi,0:\tau}(x,\epsilon^{(b)}),\quad
\mathcal S=\operatorname{Top}_{\budget}
\widehat R_\tau(h_\tau^{(1:M)},\mathcal D_t),
\label{eq:memory-seed-screen}
\end{equation}
and complete only $b\in\mathcal S$.  Memory predicts a residual over a matched
history-blind preview ranker; random-screen, wrong-memory, and $256$ shuffled-
history controls receive identical proposals, checkpoints, and NFE.  Every
$\epsilon^{(b)}$ remains an independent production-prior draw.  This restriction
is essential: fresh matched-seed continuous shaping fails to improve its cold
control (official7 $-0.000454$, CI $[-0.002209,+0.001481]$), while direct
noise-affine optimization and averaging exhibit off-manifold Goodhart behavior.
Equation~\eqref{eq:memory-seed-screen} is prospective, not a reported gain; its
frozen gates are in
\texttt{eval/PREREG\_memory\_conditioned\_seed\_screening\_20260720.md}.

\paragraph{Verifier target.}
An ideal intermediate verifier predicts terminal utility, e.g., the
Feynman--Kac potential
\begin{equation}
R_\tau^*(h,c)=\log\mathbb E\!
 [\exp\{\lambda R(z_0,c)\}\mid h_\tau=h,c],
\label{eq:terminal-value-verifier}
\end{equation}
whose small-$\lambda$ limit ranks conditional expected reward
\citep{searchdiffusion2025}.  This is the diffusion analogue of a process
verifier: its criterion is held-out pairwise ranking of final outcomes, not its
own score.  Search amplifies proxy error \citep{rewardoveropt2023,infscaling2025},
so the protocol requires a matched ordinary-preview ranker, positive held-out
alignment, history shuffles, and motion/sharpness/diversity diagnostics.

\paragraph{Why smaller late velocity is not generally better.}
For an interpolant
\begin{equation}
x_t=\alpha_t x^*+\sigma_t\epsilon,\qquad
v(x,t)=\mathbb E[\dot\alpha_t x^*+\dot\sigma_t\epsilon\mid x_t=x],
\label{eq:interpolant-velocity}
\end{equation}
the field depends on the path and schedule \citep{flowmatching2023,sit2024}.
Under a monotone time change $s=h(t)$, the same trajectory has
$\widetilde v(x,s)=v(x,t)/h'(t)$, so its norm can change without changing the
sample.  For $\alpha_t=1-t,\sigma_t=t$, the conditional target is
$\epsilon-x^*$ and need not vanish at the data endpoint.  Small raw velocity can
therefore mean a slow schedule, a static mode, or local convergence.

The preregistered diagnostic instead compares raw velocity with the
scheduler-scaled update $\|\Delta x\|/(\|x\|+\varepsilon)$ and normalized
one-step/two-half-step defect
\begin{equation}
d_{\rm step}=\frac{\|\Phi_{\Delta t}(x)-
\Phi_{\Delta t/2}(\Phi_{\Delta t/2}(x))\|}
{\|\Phi_{\Delta t}(x)-x\|+\varepsilon}.
\label{eq:step-defect}
\end{equation}
Raw velocity is promoted only if it adds multiplicity-corrected held-out
terminal-quality information beyond these schedule-aware quantities.  No such
positive claim is currently made.

\reproducibilityappendix

\section{Detailed interfaces and compute accounting}\label{app:detail-ledgers}
These ledgers collect the interfaces and costs referenced throughout the main
text; they are separated from the result tables so that accounting choices do
not interrupt the empirical narrative.
\pminterfacetable
\computeledgertable

\section{Complete PAI dimensions}\label{app:pai-full}
The main table reports aggregate official means.  The following tables expose
all eight dimensions and the cross-base ablation so that improvements cannot
hide a protected-dimension regression.
\paifulldetailtable
\paicrossbaseablationtable

\section{OpenS2V-Eval matched-pool \cvaselect{} reanalysis}\label{sec:opens2v-pilot}
Table~\ref{tab:opens2v-headline} retains every official OpenS2V axis
\citep{opens2vnexus2025}.  The internal matched pool contains four candidate
sources for each of $180$ Open-Domain tasks at fresh generation seed
$20260720$.  This is a within-pool selector evaluation, not a submission to the
public world-model leaderboard.

\begin{table}[H]
\centering
\caption{\textbf{Complete OpenS2V-Eval \cvaselect{} comparison ($B=4$; $180$ tasks).}
All seven normalized official components and their equal-weight mean are
reported; bold marks the best entry per column.  $^\dagger$Post-hoc descriptive
reanalysis: G is frozen cross-seed global utility and M is negative fresh-video
frame-difference motion; official metrics are used only after picks.}
\label{tab:opens2v-headline}
\label{tab:opens2v-pilot}
\footnotesize
\setlength{\tabcolsep}{1.2pt}
\renewcommand{\arraystretch}{1.10}
\begin{tabular}{@{}lcccccccc@{}}
\toprule
Method & Aes. & Smooth. & Motion & FaceSim & GME & Nexus & Natural & Mean \\
\midrule
\input{opens2v_table4_rows.tex}
\end{tabular}
\end{table}

\cvaselect{} selects base/Qwen/LingBot/GPT-5.4 candidates on
$32/29/63/56$ tasks.  Its mean is $0.3992$, compared with $0.4127$ for the best
fixed source; the paired category-stratified difference is $-0.0197$ with a
$95\%$ interval of $[-0.0301,-0.0096]$.  Thus the unified interface does not
transfer automatically.  The full result is bound to
\texttt{eval/artifacts/opens2v\_unified\_gm\_20260904/}
\texttt{OPEN\_S2V\_UNIFIED\_GM\_RESULT.json} (SHA256
\texttt{a89877db...8abb1}).

\section{Why the PAI memory variants were not promoted}
\label{sec:pai-developmental}
These analyses explain why model-specific fitting was not promoted into the
headline PAI result.  The historical arms are separated from the zero-memory
Table~\ref{tab:pai}.
Pairwise-rich adds four FIT-only $\theta$-PCA and four FIT-only temporal-PCA
coordinates to the Native components.  Each world model independently fits its
own PCA transforms, ridge, threshold, and train-only CERT gate; no parameters
or outcomes are shared across bases.

\begin{table}[htbp]
\centering
\caption{\textbf{PAI model-specific pretraining (official 8D).}  Each model
uses its own $174\times8$ labels and five-fold state.  All gates fail.}
\label{tab:pai-developmental}
\footnotesize
\setlength{\tabcolsep}{1.5pt}
\renewcommand{\arraystretch}{1.08}
\begin{tabular}{@{}lcccccc@{}}
\toprule
Model & no-TTS & Current PM & Pretrain & $\Delta$ [95\% CI] & CERT & Outer $+$ \\
\midrule
Wan2.2-5B & 0.759376 & 0.769758 & 0.769828 & $+.000069\ [-.000275,+.000461]$ & $1/5$ & $1/5$ \\
ABot-14B & 0.768894 & 0.772483 & 0.772483 & $0\ [0,0]$ & $0/5$ & $0/5$ \\
Cosmos3-Nano & 0.773021 & 0.776845 & 0.776704 & $-.000141\ [-.000566,+.000214]$ & $1/5$ & $0/5$ \\
\bottomrule
\end{tabular}
\end{table}

The three model files have distinct hashes.  Their scene-clustered mean effect
is $-0.000024$ (CI $[-0.000208,+0.000155]$; W/T/L $12/150/12$).
Thus model-specific fitting does not rescue the PAI quality claim.

\paragraph{Absolute-residual PM fails its gate.}
Five-fold outer cross-fitting over all $174$ scenes permits only the other four
folds to write their official seven-dimensional outcomes.  Only $1/5$ training
gates passes.  Native scores $0.739852$ and gated PM $0.739656$
($\Delta=-0.000197$, CI $[-0.000493,+0.000021]$); oracle top-1 drops from
$61/174$ to $58/174$.  Completing the eighth dimension after picks freeze gives
Native $0.769758$ and gated PM $0.769590$ ($\Delta=-0.000169$, CI
$[-0.000446,+0.000038]$).  Four fold gains are zero and the fifth is
$-0.000863$.

\paragraph{Pairwise repair remains developmental.}
The original fold-wide gate creates $170$ ties.  A margin-weighted pairwise
residual instead retains Native as a null expert and uses a train-tuned per-scene
intervention threshold.  It changes $43/174$ picks and moves the seven-dimensional
mean from $0.739852$ to $0.740252$ ($\Delta=+0.000399$, CI
$[-0.000272,+0.001126]$), with $28/131/15$ wins/ties/losses.  Only $1/5$
certification folds passes and $3/5$ outer folds are positive.  A later
train-only risk diagnostic reaches $0.770343$ ($\Delta=+0.000584$, CI
$[+0.000042,+0.001237]$) but was proposed after the outer result and therefore
is not independent confirmation.  It is superseded for model-specific inference
by the frozen protocol above, whose 5B delta is $+0.000069$ with CI
$[-0.000275,+0.000461]$.

\paragraph{Matched-history specificity and the relocation boundary.}
On $12$ task-disjoint RoboTwin tasks, DEQM changes nine scenes and obtains
$+0.006485$ with $9$ wins/$0$ losses; Candidate-DEQM changes six and obtains
$+0.006211$ with $6$ wins/$0$ losses.  Neither beats its matched shuffled-
history specificity gate, so these effects cannot be assigned independently to
memory.  The superseded residual diagnostic on all $14$ remaining relocation
tasks reaches $+0.006937$ (CI $[+0.003387,+0.011135]$), while the post-outcome
ordinary pairwise RM reaches $+0.006493$ (CI $[+0.003252,+0.010368]$); their
difference is inconclusive.  The later fresh episodes 16--19 result supersedes
this diagnostic for independent-PM inference: PM and ordinary RM are identical
on 5B and PM does not beat ordinary RM on 14B.  The residual rule's
six-intervention outer-fold PAI evaluation also fails: deployment
$-0.000222$, changed-only $-0.006433$ (CI $[-0.015561,+0.001295]$), $2$ wins/$4$
losses, and exact sign $p=0.6875$.  Together these controls identify transferable
outcome-trained reranking on this corpus, not a unique slow-memory advantage or
an independent memory confirmation.

\begin{table}[htbp]
\centering
\caption{\textbf{RoboTwin per-task gains over zero memory.}  Eight scenes/task;
RM and slow use matched outcomes, frozen 12-D features, $\alpha=100$, and
top-21 coverage.  $\Delta=\mathrm{slow}-\mathrm{RM}$.}
\label{tab:robotwin-per-task}
\footnotesize
\setlength{\tabcolsep}{3.8pt}
\begin{tabular}{lrrr}
\toprule
Task & RM & Slow & $\Delta$ \\
\midrule
move card & $+0.006631$ & $+0.006631$ & $0.000000$ \\
move stapler pad & $+0.004432$ & $+0.002843$ & $-0.001589$ \\
pick bottles & $0.000000$ & $0.000000$ & $0.000000$ \\
place bread/skillet & $+0.010142$ & $+0.010388$ & $+0.000246$ \\
place burger/fries & $0.000000$ & $0.000000$ & $0.000000$ \\
place cans/box & $0.000000$ & $0.000000$ & $0.000000$ \\
place cup & $+0.025067$ & $+0.027042$ & $+0.001975$ \\
place fan & $+0.004785$ & $+0.004785$ & $0.000000$ \\
place mouse/pad & $+0.012089$ & $+0.012089$ & $0.000000$ \\
place object/basket & $+0.002392$ & $+0.002392$ & $0.000000$ \\
place object/stand & $0.000000$ & $0.000000$ & $0.000000$ \\
stack 3 blocks & $+0.004597$ & $+0.004597$ & $0.000000$ \\
stack 2 blocks & $+0.014666$ & $+0.014666$ & $0.000000$ \\
stack 2 bowls & $+0.006097$ & $+0.011689$ & $+0.005592$ \\
\midrule
Mean ($14$ tasks) & $+0.006493$ & $+0.006937$ & $+0.000445$ \\
\bottomrule
\end{tabular}
\pmrobotvariants
\end{table}

\paragraph{Fresh-condition task-local confirmation.}
We next hold task identity fixed but move to entirely new episodes.  The
history consists of episodes $0$--$3$; evaluation uses previously unscored
episodes $8$--$11$ on the same $14$ tasks ($56$ evaluation scenes).  The
always-on task-local residual changes $28/56$ picks relative to the global model
fit on the same history and improves its mean by
$+0.001514$, but the task-bootstrap CI $[-0.002031,+0.005258]$ crosses zero;
scene W/T/L is $16/28/12$, only $6/14$ task means are positive, and the worst
scene is $-0.025177$.  Correct identity does beat all $256$ wrong-task
derangements (empirical $p=0.003891$) and raises oracle hits by four, showing a
real task-specific signal.  It nevertheless scores $-0.001191$ versus the
parent pairwise proposer without current history (CI
$[-0.006448,+0.003100]$).
The preregistered test therefore rejects the claim: four prior outcomes expose
task identity but do not yield a robust or uniformly safe PM deployment rule.

\section{Complete-budget accounting checklist}
We close the experimental appendix by restating the accounting invariant used
to compare every route.
Every quantitative comparison satisfies one of the following:
\begin{itemize}
\item no-TTS generates and scores one candidate ($\budget=1$);
\item a best-of-$\budget$ row generates and scores exactly $\budget$ candidates;
\item an oracle uses the same $\budget$ candidates but selects with the official
metric and is explicitly evaluation-only.
\end{itemize}
No result generates more candidates than its displayed budget, and no reported
method ranks a generated set larger than $\budget$ before scoring a subset.

\section{Additional PAI held-out results}
As a final robustness check, we report the held-out ordering on the scenes not
used for the earlier branch-development split.
On the last $94$ sorted PAI scenes, the unified-$\budget=8$ means are
$0.7530/0.7551/0.7566/0.7607/0.7615$ for no-TTS, Proprio, VideoReward,
the zero-memory ensemble without i2v, and the complete zero-memory ensemble.  These scenes are
disjoint from the first $80$ used for the earlier branch-development split.  The
held-out ordering matches the all-scene result and keeps the complete ensemble
above every baseline.

\section{Why separate verification counts are omitted}
The single-budget convention follows from end-to-end cost rather than
notation.  Separating generated and verified counts can suggest a saving even
when the dominant generation cost is unchanged.
For a world-model generation cost $C_{\mathrm{gen}}$ and verifier cost
$C_{\mathrm{ver}}$, a procedure that generates $G$ candidates but scores only
$V$ costs
\[
G C_{\mathrm{gen}}+V C_{\mathrm{ver}}.
\]
When video generation dominates, reducing $V$ while holding $G$ fixed is not a
meaningful claim about end-to-end test-time scaling.  Our protocol sets
$G=V=\budget$ and reports only this complete budget.  Experiments that use a
larger generated collection to choose a smaller scored subset remain useful
mechanism probes, but they are deliberately excluded from the reported method and
headline tables.


\end{document}

%% file: arxiv_title_setup.tex
\iclrfinalcopy
\makeatletter
\def\@maketitle{%
  \vbox{\hsize\textwidth
    \centering
    {\LARGE\scshape \@title\par}
    \vskip 0.18in
    {\normalsize\normalfont \@author\par}
    \vskip 0.20in
  }%
}
\makeatother
\hypersetup{
  pdftitle={From Sampling Headroom to Selection Gain: Verifier-Limited Test-Time Scaling for Video World Models},
  pdfauthor={Yuhua Jiang, Junjie Lu, Feifei Gao}
}

%% file: math_commands.tex
\newcommand{\budget}{B}                   

\newcommand{\argmin}{\operatorname*{arg\,min}}
\newcommand{\argmax}{\operatorname*{arg\,max}}

%% file: fig_teaser_main.tex
\begin{figure}[t]
\centering
\setlength{\abovecaptionskip}{2pt}
\setlength{\belowcaptionskip}{0pt}
\resizebox{0.92\linewidth}{!}{%
\begin{tikzpicture}[font=\small,>=Latex,
  box/.style={rounded corners=2.5pt, align=center, inner sep=4pt,
             minimum height=13mm, minimum width=15mm, line width=0.6pt},
  froz/.style={box, draw=cFroz!75, fill=cFroz!9},
  ver/.style={box, draw=cVer!80, fill=cVer!9},
  pickb/.style={box, draw=cPick!80, fill=cPick!10},
      gate/.style={rounded corners=2.5pt, align=center, inner sep=3pt,
                                          minimum height=14mm, text width=27mm, line width=0.7pt},
      decision/.style={rounded corners=2.5pt, align=center, inner sep=3pt,
                                                       minimum height=14mm, text width=17mm, line width=0.8pt},
  clab/.style={font=\footnotesize, text=black!70, align=center},
  it/.style={font=\footnotesize\itshape, text=black!58, align=center},
  flow/.style={-{Latex[length=2.4mm]}, line width=0.85pt, draw=black!55},
  writeA/.style={-{Latex[length=2.6mm]}, line width=1.0pt, draw=cCorr},
  warmA/.style={-{Latex[length=2.6mm]}, line width=1.0pt, draw=cMem!92!black},
]
\tikzset{
  snow/.pic={
    \foreach \a in {30,90,150}{\draw[cFroz,line width=0.6pt] (\a:2.6mm)--(\a:-2.6mm);}
            \foreach \a in {30,90,150,210,270,330}{
      \draw[cFroz,line width=0.6pt] (\a:2.6mm)--($(\a:2.6mm)+(\a+148:1mm)$);
      \draw[cFroz,line width=0.6pt] (\a:2.6mm)--($(\a:2.6mm)+(\a-148:1mm)$);}
  },
  scene/.pic={
    \draw[black!40,rounded corners=1.2pt,fill=cFroz!6] (-5mm,-5mm) rectangle (5mm,5mm);
    \fill[cFroz!13] (-4.6mm,-4.6mm) rectangle (4.6mm,-1.7mm);
    \draw[black!28,line width=0.4pt] (-4.6mm,-1.7mm)--(4.6mm,-1.7mm);
    \fill[cPick!75] (1.4mm,1.7mm) circle (1.1mm);
  },
}
\node (cond) [minimum size=10mm] {};
\pic at (cond) {scene};
\node[clab, above=1.2mm of cond] {Input $x_i$};
\node[froz, right=10mm of cond] (wm) {\textbf{Frozen proposer}\\\textbf{/ world model}\\[0.2mm]{\scriptsize no weight update}};
\pic at ([shift={(-1mm,1mm)}]wm.north east) {snow};
\coordinate (candc) at ([xshift=15mm]wm.east);
\foreach \dx/\dy in {2.4mm/-2.4mm, 1.2mm/-1.2mm, 0mm/0mm}{
  \draw[black!40, rounded corners=1.2pt, fill=white]
    ([shift={(\dx-4.5mm,\dy-4.5mm)}]candc) rectangle ([shift={(\dx+4.5mm,\dy+4.5mm)}]candc);
}
\fill[cFroz!12] ([shift={(-4.1mm,-4.1mm)}]candc) rectangle ([shift={(4.1mm,-1.4mm)}]candc);
\draw[black!28,line width=0.4pt] ([shift={(-4.1mm,-1.4mm)}]candc)--([shift={(4.1mm,-1.4mm)}]candc);
\fill[cPick!70] ([shift={(1.3mm,1.6mm)}]candc) circle (1mm);
\node[clab, above=3.4mm of candc] {$\budget$ candidates};
\node[ver, right=13mm of candc] (ver) {Apply one frozen route\\[0.2mm]{\scriptsize score all $\budget$ candidates}\\[-0.3mm]{\scriptsize no current target labels}};
\node[pickb, right=9mm of ver] (pick) {$\displaystyle\argmax_{b\le\budget}u_b$\\[0.2mm]{\scriptsize select}};
\node[minimum size=10mm, right=10mm of pick] (sel) {};
\draw[cPick!80,line width=0.8pt,rounded corners=1.2pt,fill=cPick!8]
  ([shift={(-5mm,-5mm)}]sel) rectangle ([shift={(5mm,5mm)}]sel);
\fill[cPick!13] ([shift={(-4.6mm,-4.6mm)}]sel) rectangle ([shift={(4.6mm,-1.7mm)}]sel);
\draw[black!28,line width=0.4pt] ([shift={(-4.6mm,-1.7mm)}]sel)--([shift={(4.6mm,-1.7mm)}]sel);
\fill[cPick!85] ([shift={(1.2mm,1.6mm)}]sel) circle (1.1mm);
\node[font=\normalsize,text=cPick!85] at ([shift={(-2.6mm,2.7mm)}]sel) {$\star$};
\node[clab, above=1.2mm of sel] {selected output};
\draw[flow] (cond) -- (wm);
\draw[flow] (wm.east) -- ([xshift=-5.4mm]candc);
\draw[flow] ([xshift=5.4mm]candc) -- (ver);
\draw[flow] (ver) -- (pick);
\draw[flow] (pick) -- ([xshift=-5.4mm]sel);
\coordinate (memNW) at ([xshift=-4mm, yshift=-17mm]cond.west);
\coordinate (memSE) at ([xshift=4mm, yshift=-44mm]sel.east);
\node[draw=cCorr!80, fill=black!1, rounded corners=4pt, line width=0.9pt,
      fit=(memNW)(memSE), inner sep=0pt] (mem) {};
\node[anchor=north west, font=\small\bfseries, text=cCorr!72!black]
      at ([shift={(4.5mm,-2.2mm)}]mem.north west) {Compute-Value Audit: diagnose before allocating more inference};
\node[gate, draw=cMem!75!black, fill=cMem!8, anchor=west] (g1)
      at ([shift={(4mm,-2.4mm)}]mem.west)
      {\textbf{G1: opportunity}\\[0.4mm]{\scriptsize usable headroom over the incumbent?}};
\node[gate, draw=cFroz!75, fill=cFroz!7, right=4mm of g1] (g2)
      {\textbf{G2: state}\\[0.4mm]{\scriptsize is the incumbent trustworthy?}};
\node[gate, draw=cVer!80, fill=cVer!8, right=4mm of g2] (g3)
      {\textbf{G3: action value}\\[0.4mm]{\scriptsize does the intervention improve quality?}};
\node[gate, draw=cCorr!80, fill=cCorr!7, right=4mm of g3] (g4)
      {\textbf{G4: entry fee}\\[0.4mm]{\scriptsize does recovery beat matched uniform?}};
\node[decision, draw=cPick!85!black, fill=cPick!10, right=4mm of g4] (decision)
      {\textbf{Decision}\\[0.4mm]{\scriptsize proceed / stop}};
\draw[flow] (g1) -- (g2);
\draw[flow] (g2) -- (g3);
\draw[flow] (g3) -- (g4);
\draw[flow] (g4) -- (decision);
\begin{scope}[on background layer]
  \node[rounded corners=4pt, fill=black!3, draw=black!10, line width=0.5pt,
        fit=(cond)(wm)(ver)(pick)(sel), inner xsep=4mm, inner ysep=4mm] (bg) {};
\end{scope}
\node[anchor=south, font=\footnotesize\itshape, text=black!52] at ([yshift=1.3mm]bg.north)
      {Active protocol: one complete candidate budget $\budget$};
\end{tikzpicture}}
\caption{\textbf{Domain-general Compute-Value Audit.}
\emph{Top:} a frozen proposer exposes exactly $\budget$ candidates, and one
frozen route scores all of them.  \emph{Bottom:} \cva{} audits opportunity,
state confidence, action value, and the fee relative to matched uniform compute.
State is diagnostic only; adaptive inference is authorized only after action value
clears the fee.  Flow, Cycle, PM, and anchor--explorer are audited routes, not
separate method identities.  Benchmark-specific evidence appears later.}
\label{fig:teaser}
\end{figure}

%% file: tab_anchor_ablation_main.tex
\begin{table}[H]
\centering
\caption{\textbf{Positive control: CVA certifies value when state and action align.}
On a frozen $128$-problem PRM800K confirmation set, anchor--explorer survives
sampled-anchor and exact-token controls.  Support is sparse; full estimates and
sensitivity tests are in Appendix Table~\ref{tab:anchor-ablation}.}
\label{tab:anchor-ablation-main}
\footnotesize
\setlength{\tabcolsep}{4pt}
\renewcommand{\arraystretch}{1.02}
\begin{tabular}{@{}p{0.35\linewidth}p{0.34\linewidth}p{0.24\linewidth}@{}}
\toprule
Test & Estimate [95\% CI] & Interpretation \\
\midrule
Baseline (no-TTS; greedy) & Acc. $0.7344$ & Strong base \\
Sampled-anchor policy $-$ uniform & $+0.0358\ [+.0119,+.0646]$ & Greedy unnecessary \\
Agreement signal & Acc./value $+0.8627/+0.1981$ & State routes depth \\
Policy $-$ token uniform/random & $+0.0159/+0.0259$ & Exact-cost gain \\
\bottomrule
\end{tabular}
\end{table}

%% file: tab_video_action_ladder_main.tex
\begin{table}[H]
\centering
\caption{\textbf{Richer video states still stop at action value.}
Frozen $179$-prompt VideoPhy2 ladder.  The oracle establishes actionable
heterogeneity; every deployable state is tested out of fit.  Appendix
Section~\ref{app:cva-mechanism} reports the complete tests.}
\label{tab:video-action-ladder-main}
\footnotesize
\setlength{\tabcolsep}{3pt}
\renewcommand{\arraystretch}{1.05}
\begin{tabular}{@{}
>{\raggedright\arraybackslash}p{0.26\linewidth}
>{\raggedright\arraybackslash}p{0.30\linewidth}
>{\raggedright\arraybackslash}p{0.30\linewidth}
>{\centering\arraybackslash}p{0.08\linewidth}@{}}
\toprule
State / action & Held-out result & Matched check & Gate \\
\midrule
Target-oracle router & $+3.07$ pp advantage & Evaluation-only routing upper bound & \passmark \\
Structured-set predictor & Negligible held-out gain & Matched random assignment & \stopmark \\
Transfer / supervision & No reliable held-out gain & Refit, transfer, and action-label gates & \stopmark \\
Partial-DiT preview$^\dagger$ & Degrades quality & Candidate-specific preview and refit & \stopmark \\
\bottomrule
\end{tabular}
\vspace{-1mm}
\parbox{0.98\linewidth}{\scriptsize $^\dagger$Training-selected VideoReward control.}
\end{table}

%% file: tab_width_decomposition_main.tex
\begin{table}[H]
\centering
\caption{\textbf{Width creates headroom, not recovery.}
Physics-IQ $B:4\to16$; $192$ scenes/$64$ families, $50{,}000$ family
bootstraps.  Cells show estimate and $95\%$ CI;
$\Delta\mathrm{IQ}=\mathrm{opp.}+\mathrm{align.}$.
$^\dagger$Privileged paired future.  These are previously frozen mechanism
controls; the selected \cvaselect{} score was not evaluated at $B=16$.}
\label{tab:width-decomposition}
\footnotesize
\setlength{\tabcolsep}{2.0pt}
\renewcommand{\arraystretch}{1.08}
\begin{tabular}{@{}
>{\raggedright\arraybackslash}p{0.22\linewidth}
>{\centering\arraybackslash}p{0.15\linewidth}
>{\centering\arraybackslash}p{0.19\linewidth}
>{\centering\arraybackslash}p{0.20\linewidth}
>{\centering\arraybackslash}p{0.20\linewidth}@{}}
\toprule
Route & IQ ($4\to16$) & $\Delta$IQ & Opportunity & Alignment \\
\midrule
Baseline (no-TTS) & $25.07\to25.07$ & $0$ (ref.) & -- & -- \\
True-IQ oracle & $35.09\to44.32$ & \shortstack[c]{$\mathbf{+9.23}$\\$[+7.44,+11.14]$} & -- & -- \\
\midrule
Native cycle & $27.53\to27.79$ & \shortstack[c]{$+0.26$\\$[-1.59,+2.10]$} & \shortstack[c]{$+1.78$\\$[+0.01,+3.45]$} & \shortstack[c]{$-1.52$\\$[-3.36,+0.51]$} \\
Held-out control & $29.21\to28.60$ & \shortstack[c]{$\mathbf{-0.61}$\\$[-2.73,+1.62]$} & \shortstack[c]{$\mathbf{+2.75}$\\$[+1.66,+3.74]$} & \shortstack[c]{$\mathbf{-3.36}$\\$[-5.43,-1.24]$} \\
Paired-future bound$^\dagger$ & $32.19\to36.99$ & \shortstack[c]{$\mathbf{+4.80}$\\$[+3.12,+6.57]$} & \shortstack[c]{$+6.13$\\$[+4.64,+7.70]$} & \shortstack[c]{$-1.33$\\$[-3.01,+0.47]$} \\
\bottomrule
\end{tabular}
\end{table}

%% file: tab_physics_ablation_main.tex
\begin{table}[H]
\centering
\caption{\textbf{Complete Physics-IQ component ablation ($\budget=4$).}
All seven non-empty subsets rerank the same frozen pools; G, N, and M are
train-global outcome history, native cycle, and negative motion.  Means cover
$192$ scenes/$64$ families; \cvaselect{} uses G$+$M, is retrospectively
foregrounded, and $\Delta$
is relative to it.}
\label{tab:physiq-ablation-main}
\footnotesize
\setlength{\tabcolsep}{3.0pt}
\renewcommand{\arraystretch}{1.03}
\begin{tabular}{@{}lccccc@{}}
\toprule
Signals & Wan2.2-5B & ABot-14B & Cosmos-Nano & Avg. & $\Delta$ \\
\midrule
G & 30.59 & 44.13 & 34.94 & 36.55 & $-0.05$ \\
N & 27.53 & 41.57 & 33.70 & 34.27 & $-2.34$ \\
M & 30.33 & 43.69 & 35.10 & 36.38 & $-0.23$ \\
\midrule
G$+$N & 29.86 & 43.20 & 34.31 & 35.79 & $-0.81$ \\
\textbf{\cvaselect{}} & \textbf{30.81} & \textbf{44.01} & \textbf{34.98} & \textbf{36.60} & -- \\
N$+$M & 29.14 & 42.61 & 34.58 & 35.44 & $-1.16$ \\
G$+$N$+$M & 30.06 & 43.25 & 34.74 & 36.01 & $-0.59$ \\
\bottomrule
\end{tabular}
\end{table}

%% file: tab_cross_benchmark_main.tex
\begingroup
\newcommand{\cvapass}{\ensuremath{\checkmark}}
\newcommand{\cvapartial}{\ensuremath{\triangle}}
\newcommand{\cvastop}{\ensuremath{\times}}
\newcommand{\cvana}{\textemdash}
\setlength{\intextsep}{4pt}
\setlength{\abovecaptionskip}{2pt}
\setlength{\belowcaptionskip}{3pt}
\begin{table}[H]
\centering
\caption{\textbf{CVA gate matrix.}
$\checkmark/\triangle/\times$ denote pass/partial/stop.  In each row, the first
$\times$ marks the gate that stops the audit; rows with four $\checkmark$
entries clear it.  Effects and CIs are row-local, and no-TTS is the reference.}
\label{tab:two-gates}
\footnotesize
\setlength{\tabcolsep}{1.2pt}
\renewcommand{\arraystretch}{1.12}
\begin{tabular}{@{}
>{\raggedright\arraybackslash}p{0.14\linewidth}
>{\raggedright\arraybackslash}p{0.20\linewidth}
>{\raggedright\arraybackslash}p{0.19\linewidth}
>{\raggedright\arraybackslash}p{0.25\linewidth}
>{\raggedright\arraybackslash}p{0.18\linewidth}@{}}
\toprule
Eval. & \shortstack[l]{G1\\Opportunity} & \shortstack[l]{G2\\State} &
\shortstack[l]{G3\\Action} & \shortstack[l]{G4\\Fee} \\
\midrule
\multicolumn{5}{@{}l}{\textbf{Video models}} \\
Physics-IQ\par $192$; $4\to16$ &
\shortstack[l]{\cvapass{} $+9.23$\\$[+7.44,+11.14]$} &
\cvana &
\shortstack[l]{\cvastop{} $3/3$ CIs\\cross $0$} &
\cvana \\

VideoPhy2\par $179$; $B=8$ &
\cvapass{} $+20.67$ oracle &
\shortstack[l]{rank $.839$\\MAE $.170$} &
\shortstack[l]{\cvastop{} best $0.00$\\$[-5.59,+5.59]$} &
\cvana \\

Trajectory P\&P\par $179$; $60$ NFE &
\cvana &
uncertainty mask &
\shortstack[l]{\cvastop{} $-1.68$ pp\\$[-4.47,+1.12]$} &
\shortstack[l]{\cvastop{} $0/3$ fresh\\replicas $\geq0$} \\

Adaptive depth\par $3$ WMs &
\shortstack[l]{\cvapass{} oracle $>$ unif.\\($3/3$)} &
$12$ score rules &
\cvapartial{} $42$--$69\%$ fee &
\cvastop{} $0/3>$ unif. \\

\midrule
\multicolumn{5}{@{}l}{\textbf{Reasoning}} \\
PRM800K\par $128$ &
\cvapass{} oracle $+10.94$ &
\cvapass{} agreement &
\shortstack[l]{\cvapass{} $+3.93$ vs rand.\\$[+1.14,+6.78]$} &
\shortstack[l]{\cvapass{} $+2.64$ vs unif.\\$[+0.62,+5.14]$}$^\dagger$ \\

MMLU-Pro\par $448$ &
\cvapass{} oracle $+19.87$ &
\shortstack[l]{\cvapass{} $+15.20$\\$[+10.16,+20.20]$} &
\shortstack[l]{\cvastop{} $-2.43$\\$[-5.27,+0.33]$} &
\shortstack[l]{\cvastop{} $-0.40$ vs\\matched unif.} \\
\midrule
\multicolumn{5}{@{}p{0.96\linewidth}@{}}{$^\dagger$Sparse: $7/128$ positive.} \\
\bottomrule
\end{tabular}
\end{table}
\endgroup

%% file: tab_state_action_main.tex
\begin{table}[H]
\centering
\caption{\textbf{State confidence does not imply action value.}
MMLU-Pro, Qwen3.5-2B, $n=448$; effects are percentage-point differences with
$95\%$ bootstrap CIs.  Green is pass ($\checkmark$); red is stop ($\times$).}
\label{tab:state-action-main}
\footnotesize
\setlength{\tabcolsep}{4pt}
\renewcommand{\arraystretch}{1.04}
\begin{tabular}{@{}
>{\raggedright\arraybackslash}p{0.39\linewidth}
>{\raggedright\arraybackslash}p{0.43\linewidth}
>{\raggedright\arraybackslash}p{0.12\linewidth}@{}}
\toprule
Check & Result [95\% CI] & Gate \\
\midrule
\multicolumn{3}{@{}l}{\textbf{Preconditions}} \\
no-TTS (greedy) & $29.24\%$ accuracy & base \\
Oracle@$8$ & $49.11\%$; $+19.87$ pp headroom & \passmark \\
\midrule
\multicolumn{3}{@{}l}{\textbf{Mechanism}} \\
State: agree $-$ disagree & $\mathbf{+15.20}\ [+10.16,+20.20]$ pp & \passmark \\
Action: disagree $-$ agree & $\mathbf{-2.43}\ [-5.27,+0.33]$ pp & \stopmark \\
State $-$ action & $\mathbf{+17.63}\ [+12.68,+22.50]$ pp & distinct \\
\midrule
\multicolumn{3}{@{}l}{\textbf{Policy control}} \\
Policy $-$ matched uniform & \shortstack[l]{Candidate: $-0.40\ [-1.10,+0.28]$ pp\\Token: $-0.40\ [-1.13,+0.29]$ pp} & \stopmark \\
\bottomrule
\end{tabular}
\end{table}

%% file: tab_anchor_explore.tex
\begin{table}[H]
\centering
\caption{\textbf{Anchor--explorer clears the fee.}  Qwen3-VL-4B on $128$
disjoint PRM800K problems.  Uniform matches mean depth, and random matches
exact depths.  $95\%$ problem CIs.  $K8$ consensus/oracle:
$0.7891/0.8438$.}
\label{tab:anchor-explore}
\footnotesize
\setlength{\tabcolsep}{2pt}
\renewcommand{\arraystretch}{1.00}
\begin{tabular}{@{}>{\raggedright\arraybackslash}p{0.28\linewidth}cccc@{}}
\toprule
Policy & Depth & Acc. & $\Delta$ uniform & $\Delta$ random \\
\midrule
Baseline (no-TTS; greedy) & $1$ & $0.7344$ & -- & -- \\
Matched random & $3.6763$ & $0.7498$ & $-0.0129\ [-.0299,+.0003]$ & $0$ \\
Uniform & $3.6763$ & $0.7627$ & $0$ & $+0.0129\ [-.0003,+.0299]$ \\
Anchor--explorer & $3.6763$ & $\mathbf{0.7891}$ & $\mathbf{+0.0264}\ [+.0062,+.0514]$ & $\mathbf{+0.0393}\ [+.0114,+.0678]$ \\
\bottomrule
\end{tabular}%
\end{table}

%% file: tab_anchor_ablation.tex
\begin{table}[H]
\centering
\caption{\textbf{What makes anchor--explorer work?}  These predeclared
retrospective audits use the $128$-problem confirmation set.  Token controls
include prompt and generation.  $95\%$ problem CIs.}
\label{tab:anchor-ablation}
\footnotesize
\setlength{\tabcolsep}{2pt}
\renewcommand{\arraystretch}{1.00}
\begin{tabular}{@{}
>{\raggedright\arraybackslash}p{0.38\linewidth}
>{\raggedright\arraybackslash}p{0.37\linewidth}
>{\raggedright\arraybackslash}p{0.17\linewidth}@{}}
\toprule
Test & Estimate [95\% CI] & Takeaway \\
\midrule
Baseline (no-TTS; greedy anchor) & accuracy $0.7344$ & Strong base \\
Greedy $-$ sampled (base) & baseline $+0.0156\ [-.0234,+.0547]$ & No anchor advantage \\
Greedy $-$ sampled (policy) & policy $-0.0078\ [-.0234,.0000]$ & Greedy unnecessary \\
Sampled policy $-$ uniform & $+0.0358\ [+.0119,+.0646]$ & Gain survives \\
Agree $-$ disagree & anchor accuracy $+0.8627\ [+.7706,+.9349]$ & Confidence signal \\
Disagree $-$ agree & deep-sampling value $+0.1981\ [+.0747,+.3347]$ & Compute-value signal \\
Policy $-$ token uniform & $+0.0159\ [+.0027,+.0332]$ & Exact-cost gain \\
Policy $-$ token random & $+0.0259\ [+.0061,+.0499]$ & Assignment matters \\
Positive support & $7/128$; top-5 mass $95.4\%$ & Sparse/stratum-sensitive \\
Trim top $10\%$: policy $-$ uniform & $-0.0006\ [-.0019,.0000]$ & Broad gain vanishes \\
\bottomrule
\end{tabular}%
\end{table}

%% file: tab_entry_fee.tex
\begin{table}[H]
\centering
\caption{\textbf{Video score-only rules miss the entry fee.}  For each
generator, the best of $12$ rules is shown at $B=8$; uniform matches its mean
depth.  Red is the realized deficit; orange is the evaluation-only oracle.}
\label{tab:entry-fee}
\footnotesize
\setlength{\tabcolsep}{4pt}
\renewcommand{\arraystretch}{1.08}
\begin{tabular}{lrrrrr}
\toprule
& & \multicolumn{2}{c}{vs. random} & \multicolumn{2}{c}{vs. uniform} \\
\cmidrule(lr){3-4}\cmidrule(lr){5-6}
Generator & Pools & Fee $\varphi$ & Best & Best & Oracle \\
\midrule
VideoPhy2, 5B ($K{=}100$) & $20$ & $+0.114$ & $+0.079$ \ ($69\%$) & \textcolor{red!70!black}{$-0.035$} & \textcolor{cMem}{$\mathbf{+0.099}\ [+.032,+.167]$} \\
Wan2.1-I2V-14B & $198$ & $+1.621$ & $+0.675$ \ ($42\%$) & \textcolor{red!70!black}{$-0.946$} & \textcolor{cMem}{$\mathbf{+2.554}\ [+2.24,+2.89]$} \\
Cosmos3-Nano (PIQ) & $198$ & $+2.020$ & $+0.985$ \ ($49\%$) & \textcolor{red!70!black}{$-1.035$} & \textcolor{cMem}{$\mathbf{+3.340}\ [+2.91,+3.79]$} \\
\bottomrule
\end{tabular}%
\end{table}

%% file: opens2v_table4_rows.tex
Frozen base candidate & 0.1221 & 0.5709 & 0.2734 & 0.4954 & 0.3719 & 0.6546 & 0.1338 & 0.3746 \\
Qwen agent candidate & 0.1186 & 0.5750 & \textbf{0.3035} & 0.4957 & 0.3746 & 0.6668 & 0.1759 & 0.3872 \\
\textbf{LingBot candidate} & \textbf{0.1627} & \textbf{0.6763} & 0.1716 & \textbf{0.5429} & 0.3752 & 0.6431 & \textbf{0.3171} & \textbf{0.4127} \\
GPT-5.4 skill candidate & 0.1399 & 0.5947 & 0.1855 & 0.4844 & \textbf{0.3769} & 0.6663 & 0.2338 & 0.3831 \\
\textbf{\cvaselect{} (ours)}$^\dagger$ & 0.1542 & 0.6477 & 0.1216 & 0.5373 & 0.3664 & \textbf{0.6874} & 0.2801 & 0.3992 \\
\bottomrule